\documentclass{article}

\usepackage[final]{neurips_2024}

\usepackage{amsmath}
\usepackage{amssymb}
\usepackage{amsfonts}
\DeclareMathOperator{\argmin}{argmin}

\usepackage[utf8]{inputenc} 
\usepackage[T1]{fontenc}    
\usepackage[hidelinks]{hyperref}       
\usepackage{url}            
\usepackage{booktabs}       
\usepackage{amsfonts}       
\usepackage{nicefrac}       
\usepackage{microtype}      
\usepackage{xcolor}         
\usepackage{bbm}
\usepackage{bm}
\usepackage{geometry}
\usepackage{graphicx}
\usepackage{times}
\usepackage{tikz,pgfplots}
\usepackage{xcolor}
\usepackage{floatrow}
\usepackage{comment}
\usepackage{amsthm}
\usepackage{cleveref}

\newtheorem{theorem}{Theorem}[section]
\newtheorem{lemma}[theorem]{Lemma}
\newtheorem{proposition}[theorem]{Proposition}
\newtheorem{corollary}[theorem]{Corollary}

\newtheorem{conj}[theorem]{Conjecture}
\newtheorem{definition}{Definition}[section]

\newtheorem{remark}[theorem]{Remark}
\newtheorem{assumption}[theorem]{Assumption}
\usepackage[inline]{enumitem}
\usepackage{caption} 
\usepackage{stmaryrd}

\newcommand{\risk}{\mathcal{R}}

\newcommand\pseudoindex[1]{\llbracket #1 \rrbracket}
\newcommand{\tp}{\psi}

\newcommand{\mlc}{\boldsymbol{\Upsilon}}

\newcommand{\bbR}{\mathbb{R}}

\newcommand{\bfa}{\mathbf{a}}
\newcommand{\bfA}{\mathbf{A}}
\newcommand{\bfb}{\mathbf{b}}
\newcommand{\bfB}{\mathbf{B}}

\newcommand{\bfC}{\mathbf{C}}

\newcommand{\bfD}{\mathbf{D}}
\newcommand{\bfe}{\mathbf{e}}

\newcommand{\bfK}{\mathbf{K}}

\newcommand{\bfM}{\mathbf{M}}

\newcommand{\bfN}{\mathbf{N}}

\newcommand{\bfr}{\mathbf{r}}

\newcommand{\bfT}{\mathbf{T}}
\newcommand{\bfu}{\mathbf{u}}
\newcommand{\bfU}{\mathbf{U}}
\newcommand{\bfv}{\mathbf{v}}
\newcommand{\bfV}{\mathbf{V}}
\newcommand{\bfw}{\mathbf{w}}
\newcommand{\bfW}{\mathbf{W}}
\newcommand{\bfx}{\mathbf{x}}
\newcommand{\bfX}{\mathbf{X}}
\newcommand{\bfy}{\mathbf{y}}

\newcommand{\bfz}{\mathbf{z}}

\newcommand{\tr}{\mathsf{tr}} 
\newcommand{\vect}{\mathsf{vec}} 
\newcommand{\rank}{\mathrm{rank}}

\newcommand{\myheader}[1]{\noindent\textbf{#1}.}

\newcommand{\olbfm}{\overline{\mathbf{m}}}
\newcommand{\olbfx}{\overline{\mathbf{x}}}
\newcommand{\olbfM}{\overline{\mathbf{M}}}
\newcommand{\olbfX}{\overline{\mathbf{X}}}

\title{The Implicit Bias of Gradient Descent on Separable Multiclass Data}

\author{%
Hrithik Ravi$^{1}$ \quad Clayton Scott$^{1}$ \quad Daniel Soudry$^{2}$ \quad Yutong Wang$^{3}$\\
$^1$University of Michigan \quad $^2$Technion - Israel Institute of Technology\\ 
$^3$Illinois Institute of Technology\\
\texttt{\{hrithikr, clayscot\}@umich.edu}\\
\texttt{daniel.soudry@gmail.com}\\
\texttt{ywang562@iit.edu}
}

\begin{document}

\maketitle

\begin{abstract}
Implicit bias describes the phenomenon where optimization-based training algorithms, without explicit regularization, show a preference for simple estimators even when more complex estimators have equal objective values.
Multiple works have developed the theory of implicit bias for binary classification under the assumption that the loss satisfies an \emph{exponential tail property}. However, there is a noticeable gap in analysis for multiclass classification, with only a handful of results which themselves are restricted to the cross-entropy loss. In this work, we 
employ the framework of Permutation Equivariant and Relative Margin-based (PERM) losses \citep{wang2023unified} to introduce a multiclass extension of the exponential tail property. This class of losses includes not only cross-entropy but also other losses.
Using this framework, we extend the implicit bias result of \citet{soudry2018implicit} to multiclass classification. 
Furthermore, our proof techniques closely mirror those of the binary case, thus illustrating the power of the PERM framework for bridging the binary-multiclass gap.
\end{abstract}

\section{Introduction}
Overparameterized models such as neural networks have shown state-of-the-art performance in many applications, despite having the potential to overfit.
\citet{zhang2021understanding}
demonstrate that
this potential is indeed realizable by training real-world models to fit random noise.
In recent years, there have been several research efforts that aim to understand the impressive performance of overparametrized models despite this ability to overfit. Both the model architecture and the training algorithms for selecting the weights have been investigated in this regard.

Work on \emph{implicit bias} \citep{soudry2018implicit,ji2020gradient,vardi2022implicit} has focused on the latter factor.
Implicit bias is the hypothesis that gradient-based methods
have a built-in preference for models with low-complexity.
This hypothesis is perhaps best understood in the setting of
(unregularized) empirical risk minimization for learning a linear model under the assumption of linearly separable data. 
\citet{soudry2018implicit} showed that 
in binary classification, implicit bias holds when the loss has the exponential tail property \citep[Theorem 3]{soudry2018implicit}. The same work also demonstrated implicit bias in the multiclass setting for the cross-entropy loss, but implicit bias for a more broadly defined class of losses in the multiclass case is left open. In this work, we 
extend the notion of the exponential tail property to multiclass losses
and prove that the property is sufficient for implicit bias to occur in the multiclass setting. Toward this end, we employ the framework of 
permutation equivariant and relative margin-based (PERM) losses \citep{wang2023unified}.

\subsection{Contributions}
\paragraph{Multiclass extension of the exponential tail property (Definition~\ref{def: proposed_multiclass_exp_tail})}
It is unclear how the exponential tail property for binary margin losses should be extended to the multiclass setting.
By using the PERM framework, we provide a multiclass extension that 
generalizes the exponential tail property to multiclass (Definition~\ref{def: proposed_multiclass_exp_tail} in \Cref{section:multiclass-exponential-tail}).  We further verify that this property holds for some common losses.

\paragraph{Sufficiency of the exponential tail property for implicit bias (Theorem~\ref{main_theorem})}
We prove that the proposed multiclass exponential tail property
is sufficient for implicit bias.
More precisely, we show in Theorem~\ref{main_theorem} that for almost all linearly separable multiclass datasets, given a convex, (\(\beta\)-smooth, strictly decreasing) PERM loss 
satisfying the exponential tail property in Definition~\ref{def: proposed_multiclass_exp_tail}, gradient descent exhibits directional convergence to the hard-margin multiclass SVM.

\subsection{Related Work}

\citet{soudry2018implicit} 
show that gradient descent, applied to \emph{unregularized} empirical risk minimization, converges to the hard-margin SVM solution at a slow logarithmic rate, provided the loss satisfies the exponential tail property (defined below). \cite{nacson2019convergence} improve the convergence rate using a specific step-size schedule. \citet{ji2019implicit} extend implicit bias to the setting of \emph{quasi-complete separation} \citep{candes2020phase}, where the two classes are linearly separated but with a margin of zero.
Many works have also considered gradient-based methods beyond gradient descent. For example, \citet{gunasekar2018characterizing} examine the implicit bias effects of mirror descent \citep{beck2003mirror}, steepest descent \citep{boyd2004convex}, and \emph{adaptive} gradient descent \citep{duchi2011adaptive,kingma2015adam}.
\citet{cotter2012kernelized,clarkson2012sublinear,ji2021fast} study first order methods that are designed specifically to approach the hard-margin SVM as quickly as possible. 

Results for the multiclass setting are more scarce, and are \emph{always} specific to cross-entropy. 
\citet{soudry2018implicit} establish implicit bias for cross-entropy loss. \citet{lyu2019gradient} focus on homogeneous predictors and prove convergence of GD on cross-entropy loss to a KKT point of the margin-maximization problem.
\cite{lyu2019gradient} proves convergence of  gradient flow to a generalized
 max-margin classifier for multiclass classification with cross-entropy loss using homogeneous models.\footnote{\citet{lyu2019gradient} could be thought of as analyzing losses beyond CE, but the optimization problem would be non-convex so convergence might not be to a global minimum. See Appendix \ref{appendix: discussion_of_lyu_li} for a more detailed discussion.} In the special case when the model are linear classifiers, the generalized max-margin classifier reduces to the classical hard-margin SVM.
\citet{lyu2021gradient} consider two-layer neural networks and prove convergence of GD on cross-entropy loss to the max-margin solution under an additional assumption on the data, that both \(\bfx\) and its negative counterpart \(-\bfx\) must belong to the dataset. 
\citet{wang2023benign} prove that in certain overparameterized regimes, gradient descent on squared loss leads to an equivalent solution to gradient descent on cross-entropy loss.

Beyond work establishing (rate of) convergence to the max-margin classifier, there is also a separate line of work \citep{shamir2021gradient,schliserman2022stability,schliserman2023tight} focusing on the \emph{generalization} aspect of implicit bias. These works examine the binary classification setting, with the exception of \citet{schliserman2022stability} who consider cross-entropy.

\subsection{Notations}

Let \(K \ge 2\) and \(d \ge 1\) denote the number of classes and feature space dimension, respectively. Let \([K] := \{ 1,2,\dots, K\}\). Vectors are denoted by boldface lowercase letters, e.g., \(\mathbf{v} \in \mathbb{R}^K\) whose entries are denoted by \(v_j\) for \(j \in [K]\). Likewise, matrices are denoted 
by boldface uppercase letters, e.g., \(\mathbf{W} \in \mathbb{R}^{d \times K}\). The columns of \(\mathbf{W}\) are denoted \(\mathbf{w}_1,\dots, \mathbf{w}_K\). By \(\mathbf{0}_{n}\) and \(\mathbf{1}_{n}\) we denote the \(n\)-dimensional vectors of all 0's and all 1's respectively. The \(n \times n\) identity matrix is denoted by \(\mathbf{I}_n\).

By \(\|\mathbf{v}\|\) we denote the Euclidean norm of vector \(\mathbf{v}\). \(\|\mathbf{A}\|_{2}\) is the spectral norm of matrix \(\mathbf{A}\). Given two vectors \(\mathbf{w}, \mathbf{v} \in \mathbb{R}^{k}\), we write \(\mathbf{w} \succeq \mathbf{v}\) (resp. \(\mathbf{w} \succ \mathbf{v}\)) if \(w_j \geq v_j\) (resp. \(w_j > v_j\)) for all \(j \in [k]\); similarly we write \(\mathbf{w} \preceq \mathbf{v}\) (resp. \(\mathbf{w} \prec \mathbf{v}\)) if \(w_j \leq v_j\) (resp. \(w_j < v_j\)) for all \(j \in [k]\). On the other hand, if \(\mathbf{A}\) and \(\mathbf{B}\) are equally-sized \emph{symmetric matrices}, then by \(\mathbf{A} \succeq \mathbf{B}\) (resp. \(\mathbf{A} \preceq \mathbf{B}\)) we mean that \(\mathbf{A} - \mathbf{B}\) (resp. \(\mathbf{B} - \mathbf{A}\)) is positive semi-definite, i.e. \(\mathbf{A} - \mathbf{B} \succeq 0\) (resp. \(\mathbf{B} - \mathbf{A} \succeq 0\)).

A bijection from \([k]\) to itself is called a permutation on \([k]\). Denote by \(\mathtt{Sym}(k)\) the set of all permutations on \([k]\). For each \(\sigma \in \mathtt{Sym}(k)\), let \(\mathbf{S}_{\sigma}\) denote the permutation matrix corresponding to \(\sigma\). In other words, if \(\mathbf{v} \in \mathbb{R}^k\) is a vector, then \([\mathbf{S}_{\sigma}\mathbf{v}]_j 
= \mathbf{v}_{\sigma(j)}\).

\section{Multiclass Loss Functions}

In multiclass classification, a classifier is typically represented in terms of a \emph{class-score function} \(f  = (f_1,\dots, f_K): \mathbb{R}^d \to \mathbb{R}^K\), which maps an input \(\mathbf{x} \in \mathbb{R}^d\) to a vector \(\mathbf{v} := f(\mathbf{x})\) of class scores.
For instance, \(f\) may be a feed-forward neural network and \(\mathbf{v}\) in this context is sometimes referred to as the logits.
The label set is $[K]$, and a label is predicted as \(\mathrm{argmax}_{j} f_j(\mathbf{x})\). 
A \(K\)-ary multiclass loss function
is a vector-valued function \(\mathcal{L} = (\mathcal{L}_1, \dots,\mathcal{L}_K) : \mathbb{R}^K \rightarrow \mathbb{R}^K\) where \(\mathcal{L}_y(f(\mathbf{x}))\) is the loss incurred for outputting \(f(\mathbf{x})\) when the ground truth label is \(y\).

In binary classification, a classifier is typically represented using a function \(g :  \mathbb{R}^d \to \mathbb{R}\). The label set is $\{-1,1\}$, and labels are predicted as \(\mathbf{x} \mapsto \mathrm{sign}(g(\mathbf{x}))\).
A \emph{binary margin loss} is a function of the form \(\psi : \mathbb{R} \to \mathbb{R}\) where 
\(\psi(y g(\mathbf{x}))\) is the loss incurred for outputting \(g(\mathbf{x})\) when the ground truth label is \(y\). 
Margin losses have been central to the development of the theory of binary classification, and the lack of a multiclass counterpart to binary margin losses may have impaired the development of corresponding theory for multiclass classification.  To address this issue, \citet{wang2023unified} introduce PERM losses as a bridge between binary and multiclass classification.

\subsection{Permutation equivariant and relative margin-based (PERM) losses}





Assume the label set is $[K]$.  Define
\footnote{
Also see \cite[Definition~2]{wang2023unified}.} 
the matrix \(\mathbf{D} := [-\mathbf{I}_{K-1} \quad \mathbf{1}_{K-1}] \in \mathbb{R}^{(K-1) \times K}\). Observe that 
\(\mathbf{D} \mathbf{v} = (v_K-v_1,v_K-v_2,\dots, v_K - v_{K-1})^\top\) for all \(\mathbf{v} \in \mathbb{R}^K\). 


\begin{definition}[PERM loss \citep{wang2023unified}]\label{def: perm_loss}
Let \(K \geq 2\) be an integer, and \(\mathcal{L}\) be a \(K\)-ary multiclass loss function. We say that \(\mathcal{L}\) is
\begin{enumerate}
    \item \emph{permutation equivariant} if \(\mathcal{L}(\mathbf{S}_\sigma\mathbf{v}) = \mathbf{S}_\sigma\mathcal{L}(\mathbf{v})\) for all \(\mathbf{v} \in \mathbb{R}^K\) and \(\sigma \in \mathtt{Sym}(K)\),
    \item \emph{relative margin-based} if for each \(y \in [K]\) there exists a function \(\ell_y : \mathbb{R}^{K-1} \rightarrow \mathbb{R}\) so that \(\mathcal{L}_y(\mathbf{v}) = \ell_y(\mathbf{D}\mathbf{v}) = \ell_y(v_K - v_1, v_K - v_2, \dots, v_K - v_{K-1)}\), for all \(\mathbf{v} \in \mathbb{R}^K\). We refer to the vector-valued function \(\ell := (\ell_1, \dots , \ell_K)\) as the \emph{reduced form} of \(\mathcal{L}\).
    \item \emph{PERM} if \(\mathcal{L}\) is both \underline
    {p}ermutation \underline
    {e}quivariant and \underline
    {r}elative \underline
    {m}argin-based. In this case, the function \(\psi := \ell_K\) is referred to as the \emph{template} of \(\mathcal{L}.\)
\end{enumerate}
\end{definition}

\citet{wang2023unified} show that PERM losses are characterized by their template $\psi$. To show this, they introduce the \emph{matrix label code}, an encoding of labels as matrices. Thus, for each \(y \in [K-1]\), 
let \(\mlc_y\) be the $(K-1) \times (K-1)$ identity matrix, but with the \(y\)-th column replaced by all \(-1\)'s.
For \(y =K\), let \(\mlc_y\) be the identity matrix.  
Note that when \(K=2\), this definition reduces to \(\mlc_y = (-1)^y\), the standard encoding of labels in the binary setting. Observe that (after permutation) $\mlc_y\mathbf{D}\mathbf{v} = (v_y-v_1,v_y-v_2,\dots, v_y - v_{K})^\top \in \mathbb{R}^{K-1}$, where the $v_y - v_y = 0$ entry is omitted. Please see \citet[Lemma B.2]{wang2023unified} for a simple proof.


\begin{theorem}[\citet{wang2023unified}]\label{theorem:relative-margin-form}
Let \(\mathcal{L} : \mathbb{R}^K \rightarrow \mathbb{R}^K\) be a PERM loss with template \(\psi\), and let \(v \in \mathbb{R}^K\) and \(y \in [K]\) be arbitrary. Then \(\psi\) is a symmetric function. Moreover, 
\begin{equation}
    \mathcal{L}_y\left(\mathbf{v}\right) = \psi\left(\mlc_{y}\mathbf{D}\mathbf{v}\right). \label{eq: relative-margin-form}
\end{equation}
Conversely, let \(\psi : \mathbb{R}^{K-1} \rightarrow \mathbb{R}\) be a symmetric function. Define a multiclass loss function \(\mathcal{L} = (\mathcal{L}_1, \dots , \mathcal{L}_k) : \mathbb{R}^K \rightarrow \mathbb{R}^K\) according to Eqn. \eqref{eq: relative-margin-form}. Then \(\mathcal{L}\) is a PERM loss with template \(\psi\).
\end{theorem}

Theorem \ref{theorem:relative-margin-form} shows that a PERM loss is characterized by its template $\psi$.
The right hand side of Eqn. \eqref{eq: relative-margin-form} is referred to as the \emph{relative margin form} of the loss, which extends binary margin losses to multiclass. As noted by \citet{wang2023unified}, an advantage of the relative margin form is that it decouples the labels from the predicted scores, which facilitates analysis. Our results below support this understanding.

Many losses in the literature are PERM losses, including the cross-entropy loss whose template is
\(\psi(\mathbf{u}) = \log(1 + \sum_{i=1}^{K-1}\exp(-u_i))\), 
the 
multiclass exponential loss \citep{mukherjee2013theory} whose template is
\(\psi(\mathbf{u}) =  \sum_{i=1}^{K-1}\exp(-u_i)\), 
and the PairLogLoss \citep{wang2022rank4class} whose template is = \(\psi(\mathbf{u}) = \sum_{i=1}^{K-1} \log(1 + \exp(-u_i))\). See \citet{wang2023unified} for other examples.

\subsection{Regularity assumptions on loss functions}

Let \(\mathcal{L}\) be a PERM loss with differentiable template \(\psi\). If 
\begin{align*}
\frac{\partial \psi}{\partial u_i}\left(\mathbf{u}\right) < 0, \quad \mbox{ for all \(i \in \{1,2\dots, K-1\}\), \(\mathbf{u} \in \mathbb{R}^{K-1}\),}
\end{align*}
i.e.,
the gradient of the template
is entrywise strictly negative, then we say that the PERM loss \(\mathcal{L}\) is \emph{strictly decreasing}. In this case, we write $\nabla\psi \prec \mathbf{0}$, where $\mathbf{0}$ is the 0-vector. 
If the template is differentiable, then it is convex if:
\begin{align*}
   \psi(\mathbf{u_1}) \geq \psi(\mathbf{u_2}) + \nabla\psi(\mathbf{u_2})^{\top}(\mathbf{u_1}-\mathbf{u_2}), \quad \mbox{ for all \(\mathbf{u_1}, \mathbf{u_2} \in \mathbb{R}^{K-1}\).} 
\end{align*}
If \(\psi\) is twice-differentiable, this is equivalent to saying that the Hessian is positive-semidefinite:
\begin{align*}
\nabla^{2}\psi\left(\mathbf{u}\right) \succeq 0 \quad \mbox{ for all \(\mathbf{u} \in \mathbb{R}^{K-1}\).}
\end{align*}
Finally, the template is said to be $\beta$-smooth if its gradient is $\beta$-Lipschitz:
\begin{align*}
    \|\nabla\psi\left(\mathbf{u_1}\right)-\nabla\psi\left(\mathbf{u_2}\right)\| \leq \beta\|\mathbf{u_1}-\mathbf{u_2}\|, \quad \mbox{ for all \(\mathbf{u_1}, \mathbf{u_2} \in \mathbb{R}^{K-1}\).}
\end{align*}
If $\psi$ is twice-differentiable, this is equivalent to saying that the maximum eigenvalue of its Hessian is bounded by $\beta$:
\begin{align*}
\|\nabla^{2}\psi\left(\mathbf{u}\right)\|_{2} \leq \beta \quad \mbox{ for all \(\mathbf{u} \in \mathbb{R}^{K-1}\),}
\end{align*}
where \(\|\mathbf{A}\|_{2}\) is the spectral norm of matrix \(\mathbf{A}\).

\subsection{Multiclass analogue of exponential tail property}\label{section:multiclass-exponential-tail}

\begin{figure}
\centering
\includegraphics[width=1.0\linewidth]{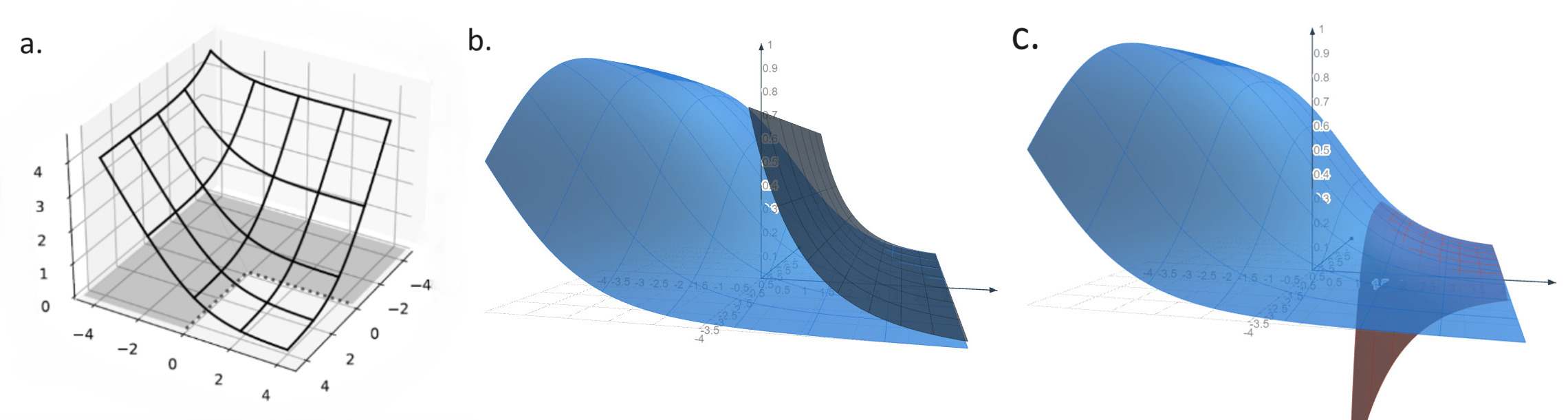}
\caption{\label{fig:multiclass_CE_exp_tail} An illustration of the exponential tail property for the cross entropy/multinomial logistic loss when \(K=3\). \emph{Panel a.} Plot of \(\psi(\mathbf{u})=\log(1+ \exp(-u_1) + \exp(-u_2))\), the template for the multinomial logistic loss. Note that the complement of the positive orthant in the domain \(\mathbb{R}^2\) is shown in gray.
\emph{Panel b. and c.} Plot of the upper bound (shown in black) and lower bounds (red) of
\(-\frac{\partial \psi}{\partial u_1}\) (blue)
respectively.
These bounds are from Appendix~\ref{appendix: CE_exp_tail}
where \(u_\pm = 0\) and \(c=1\).
Note that the lower bound is valid in the positive orthant, i.e., the red surface is below the blue one there.}\label{figure:multiclass_CE_exp_tail}
\end{figure}

In the binary setting, the exponential tail property defined in prior work (\citet{soudry2018implicit}, \citet{nacson2019convergence}, \citet{ji2020gradient}) is assumed to hold for the negative \emph{derivative} of the loss. Similarly, in the multiclass setting we are interested in bounding the negative \emph{gradient} of the PERM loss template.

\begin{definition}[Multiclass exponential tail property]
  \label{def: proposed_multiclass_exp_tail}
     A multiclass PERM loss with template $\psi: \mathbb{R}^{K-1} \rightarrow \mathbb{R}$ has the \emph{exponential tail (ET)} property if there exist $u_{+}, {u}_{-} \in \mathbb{R}$ and positive $c > 0$ such that
     for all \(i \in [K-1]\) the following holds:
\begin{align*}
\forall \mathbf{u} \mbox{ s.t. } \min_{j \in [K-1]} u_j > u_{+},  \mbox{ we have } &-\frac{\partial \psi}{\partial {u}_{i}}\left(\mathbf{u}\right) \leq c\exp(-u_{i}), \quad \mbox{and}\\
\forall \mathbf{u} \mbox{ s.t. } \min_{j \in [K-1]} u_j > u_{-}, \mbox{ we have }
&-\frac{\partial \psi}{\partial {u}_{i}}\left(\mathbf{u}\right) \geq c\left(1-\sum_{j \in [K-1]}\exp\left(-u_{j}\right)\right)\exp(-u_{i}).
\end{align*}
\end{definition}



\begin{remark}\label{remark: CE_and_exp_have_exp_tail}  We show in Appendix \ref{appendix: losses_that_satisfy_assumptions} that cross-entropy (CE), multiclass exponential loss, and PairLogLoss all have this property.
\end{remark}

\section{Main Result}\label{section:main-result}
Consider a dataset $\left\{ (\mathbf{x}_{n},y_{n})\right\}_{n=1}^{N}$,
with $\mathbf{x}_n\in\mathbb{R}^d$ and class labels $y_{n}\in[K]:=\left\{ 1,\dots,K\right\}$. 
The class score function for class $k$ is $f_k(\mathbf{x}) = \mathbf{w}_k^T \mathbf{x}$.
Define $\mathbf{X} \in \mathbb{R}^{d \times N}$ to be the matrix whose $n$th column is $\mathbf{x}_{n}$. Define $\mathbf{W} \in \mathbb{R}^{d \times K}$ to be the matrix whose $k$th column is 
$\mathbf{w}_{k}$. 
The learning objective is
\begin{equation}
\mathcal{R}\left(\mathbf{W}\right)=\sum_{n=1}^{N}\mathcal{L}_{y_{n}}\left(\mathbf{W}^{\top}\mathbf{x}_{n}\right) \,.\label{eq: general_risk}
\end{equation}

From Eqn. \ref{eq: relative-margin-form}, if $\mathcal{L}$ is a PERM loss, then $\mathcal{L}_y\left(\mathbf{v}\right) = \psi\left(\mlc_y \mathbf{D} \mathbf{v}\right)$, and the learning objective becomes
\begin{align}
\mathcal{R}(\mathbf{W})
= \sum_{i=1}^{N}\psi\left(\mlc_{y_{i}}\mathbf{D}\mathbf{W}^{\top}\mathbf{\mathbf{x}}_{i}\right) \,. \label{eq: risk_perm_form}
\end{align}
Up to permuting the entries, $\mlc_{y_{i}}\mathbf{D}\mathbf{W}^{\top}{\mathbf{x}_{i}}$ is equal to the $\left(K-1\right)$-dimensional vector of relative-margins 
\(
\left[
(\mathbf{w}_{y_i}- \mathbf{w}_{1})^{\top}\mathbf{x}_i, \, (\mathbf{w}_{y_i} - \mathbf{w}_{2})^{\top}\mathbf{x}_i,\, \dots , \,(\mathbf{w}_{y_i} - \mathbf{w}_{K})^{\top}\mathbf{x}_i
\right]^{\top}\), where the \(0\)-valued entry 
$(\mathbf{w}_{y_i} - \mathbf{w}_{y_i})^{\top}\mathbf{x}_i$  is omitted. This follows from \citet[Lemma B.2]{wang2023unified}.

We are now ready to state our assumptions on the loss:
\begin{assumption}\label{assumption: loss_smoothness_dec_nonneg}
The PERM loss's template \(\psi\) is convex, \(\beta\)-smooth, strictly decreasing and non-negative.
\footnote{
We note that in the binary case 
the implicit bias result in \cite[Theorem 3]{soudry2018implicit} does not require the loss to be convex. Closing this binary-multiclass gap is an open question.}
\footnote{Note that multiclass exponential loss \(\psi\left(\mathbf{u}\right)\) does not have a global smoothness constant. However, we show in Appendix \ref{appendix: exp_loss_pseudo_smoothness} that any learning rate \(\eta < 1/\left(B^{2}\mathcal{R}\left(\mathbf{w}(0)\right)\right)\) is sufficient for the gradient descent iterates to achieve local smoothness, where \(B = \sqrt{\left(2K-2\right)} \sum_{i=1}^{N}\|\bfx_i\|\).}
\end{assumption}

\begin{assumption}\label{assumption: loss_exp_tail}
    The PERM loss has exponential tail as defined in Definition \ref{def: proposed_multiclass_exp_tail}.
\end{assumption}
 
To optimize Eqn. \eqref{eq: general_risk} we employ gradient descent with fixed learning rate $\eta$. Define $\bfw := \vect(\bfW)$ where $\vect$ denotes  vectorization by column-stacking (See \Cref{definition:calculus:vectorization}), and let the gradient descent iterate at time $t$ be $\mathbf{w}\left(t\right)$. 
Then:
\begin{align*}
\mathbf{w}\left(t+1\right)=\mathbf{w}\left(t\right)-\eta\nabla\risk\left(\mathbf{w}(t)\right).
\end{align*}
Define the ``matrix-version'' of the trajectory \(\bfW(t) \in \bbR^{d \times K}\) such that \(\bfw(t) = \vect(\bfW(t))\).
Throughout this work, we frequently work with the risk as a \emph{matrix}-input scalar-output function \(\risk(\bfW)\),  and
as a \emph{vector}-input scalar-output function \(\risk(\bfw)\).

These two formulations will each be useful in different situations. For instances, adopting the matrix perspective can facilitate calculation of bounds, e.g., in \Cref{section:bounding-the-second-term}. 
On the other hand, the vectorized formulation is easier for defining the Hessian of the risk \(\nabla^2 \risk(\bfw)\). See \Cref{appendix: matrix_calculus} for detail.

We focus on linearly separable datasets:
\begin{assumption}\label{assumption: dataset_linearly_separable}
The dataset is linearly separable, i.e. there exists $\mathbf{w} \in \mathbb{R}^{dK}$ such that $\forall n \in [N],\forall k \in [K] \backslash \{y_n\} : \mathbf{w}_{y_n}^\top\mathbf{x}_n \ge \mathbf{w}_k^\top\mathbf{x}_n + 1$. Equivalently, there exists $\mathbf{W} \in \mathbb{R}^{d \times K}$ such that $\forall n \in [N]$, $\mlc_{y_{n}}\mathbf{D}\mathbf{W}^{\top}\mathbf{x}_{n} \succeq \mathbf{1}.$
\end{assumption}

Finally, let $\mathbf{\hat{w}}$ be the multiclass hard-margin SVM solution for the linearly separable dataset:
\begin{equation}
\label{eq: multiclass_SVM}
\hat{\mathbf{w}} = \argmin_{\mathbf{w}} \mbox{$\frac12$} \Vert \mathbf{w}\Vert^2 \,\text{ s.t.}\,\forall n,\forall k \ne y_n : \mathbf{w}_{y_n}^\top\mathbf{x}_n \ge \mathbf{w}_k^\top\mathbf{x}_n + 1.
\end{equation}

Now we state the main result of the paper:

\begin{theorem}
    \label{main_theorem} For any PERM loss satisfying Assumptions \ref{assumption: loss_smoothness_dec_nonneg} and \ref{assumption: loss_exp_tail}, for all linearly separable datasets such that 
Assumption~\ref{assumption:w_tilde_existence} holds, any sufficiently small learning rate $0 < \eta < 2\beta^{-1}\sigma^{-2}_{\text{max}}\left(\mathbf{X}\right)$, and any initialization $\mathbf{w}(0)$, the iterates of gradient descent will behave as
    \begin{align*}
        \mathbf{w}(t) = \hat{\mathbf{w}}\log(t) + \boldsymbol\rho(t)
    \end{align*}
    where the norm of the residual,  $\Vert\boldsymbol\rho(t)\Vert$, is bounded. This implies a directional convergence behavior:
    \begin{align*}  \lim_{t\rightarrow\infty}\frac{\mathbf{w}\left(t\right)}{\left\Vert \mathbf{w}\left(t\right)\right\Vert }=\frac{\hat{\mathbf{w}}}{\left\Vert \hat{\mathbf{w}}\right\Vert}.
    \end{align*}
\end{theorem}

In Appendix~\ref{section:experiments}, we show experimental results demonstrating implicit bias towards the hard margin SVM when using the PairLogLoss, in line with
    \Cref{main_theorem}.


\section{Proof Sketch}\label{section:proof-sketch}

In this section we will overview the proof of the result. Along the way, we prove lemmas that extend to the multiclass setting results from \citet{soudry2018implicit}. The extensions are facilitated by the PERM framework, in particular the relative margin from of the loss.

We adopt the notation of \cite{soudry2018implicit} where possible throughout this proof. Recalling the notation and definitions from the paper: let us define the standard basis $\mathbf{e}_{k}\in\mathbb{R}^{K}$ such that $\left(\mathbf{e}_{k}\right)_{i}=\delta_{ki}$ (where \(\delta\) is the Kronecker-delta function), and the $d$-dimension identity matrix  $\mathbf{I}_{d}$. Define $\mathbf{A}_{k}\in\mathbb{R}^{dK\times d}$ as the Kronecker product between $\mathbf{e}_{k}$ and $\mathbf{I}_{d}$, i.e.
$\mathbf{A}_{k}=\mathbf{e}_{k}\otimes\mathbf{I}_{d}$. We can then relate the original $k^{th}$-class predictor $\mathbf{w}_k$ to the long column-vector $\mathbf{w}$ as follows:  $\mathbf{A}_{k}^{\top}\mathbf{w}=\mathbf{w}_{k}$. Next define $\tilde{\mathbf{x}}_{n,k} :=(\mathbf{A}_{y_n}-\mathbf{A}_k)\mathbf{x}_n$. Using this notation, the multiclass SVM  becomes
\begin{equation}
   \argmin_{\mathbf{w}} \mbox{$\frac12$} \Vert\mathbf{w}\Vert^2 \quad \text{ s.t.}\quad \forall n,\forall k \ne y_n: \mathbf{w}^\top \tilde{\mathbf{x}}_{n,k} \ge 1  \label{eq: rewritten_multiclass_SVM_xtilde}
\end{equation}
For each $k \in [K]$, define $\mathcal{S}_k=\arg \min_n(\hat{\mathbf{w}}_{y_n}-\hat{\mathbf{w}}_k)^\top\mathbf{x}_n 
=
\{n :(\hat{\mathbf{w}}_{y_n}-\hat{\mathbf{w}}_k)^\top\mathbf{x}_n  =1\}
$, i.e., the $k^{th}$ class support vectors. From the KKT optimality conditions for Eqn. \eqref{eq: rewritten_multiclass_SVM_xtilde}, we have for some dual variables $\alpha_{n,k} > 0$ that
\begin{equation} \label{eq: multiclass_SVM_coefficients}
\hat{\mathbf{w}} = \sum_{n=1}^{N} \sum_{k=1}^{K} \alpha_{n,k} \tilde{\mathbf{x}}_{n,k} \mathbbm{1}_{n\in \mathcal{S}_k}.
\end{equation}
Finally, define 
\begin{equation}
    \label{equation:residual}
\mathbf{r}\left(t\right)
=\mathbf{w}\left(t\right)-\log\left(t\right)\hat{\mathbf{w}}-\tilde{\mathbf{w}}
\end{equation}
where $\tilde{\mathbf{w}}$ is a solution to
\begin{equation}
\label{equation:w-tilde-defining-condition}
\forall k \in [K], \forall n \in \mathcal{S}_k:\,\eta\exp\left(-\mathbf{x}_{n}^{\top}\left(\tilde{\mathbf{w}}_{y_n}-\tilde{\mathbf{w}}_{k}\right)\right)=\alpha_{n, k}.
\end{equation}
In \cite{soudry2018implicit}, the existence of  $\tilde{\mathbf{w}}$ is proven for the binary case for almost all datasets, and assumed in the multiclass case.
Here, we also state the existence of \(\tilde{\mathbf{w}}\) as an additional assumption:
\begin{assumption}\label{assumption:w_tilde_existence}
Eqn.~\ref{equation:w-tilde-defining-condition}
 has a solution, denoted \(\tilde{\mathbf{w}}\).
\end{assumption}
We pose the problem of proving Assumption~\ref{assumption:w_tilde_existence} for almost all datasets as a conjecture
in 
Appendix~\ref{section:wtildeconjecture}, where we also show experimentally that on a large number (100 instances for each choice of \(d \in \{2,3,4, 5,6\}\) and \(K \in \{3,4,5,6\}\)) of synthetically generated linearly separable datasets, Assumption~\ref{assumption:w_tilde_existence} indeed holds.

Note that $\mathbf{r}(t) = \boldsymbol\rho(t) - \tilde{\mathbf{w}}$, and $\tilde{\mathbf{w}}$ is independent of $t$, so bounding $\mathbf{r}(t)$ is equivalent to bounding $\boldsymbol\rho(t)$. 
Following the same steps as \citet[Appendix E.3]{soudry2018implicit}:
\begin{equation} \label{eq: r(t+1) norm}
  \left\Vert \mathbf{r}\left(t+1\right)\right\Vert ^{2}
-
\left\Vert \mathbf{r}\left(t\right)\right\Vert^{2}
  =
  \underbrace{\left\Vert \mathbf{r}\left(t+1\right)-\mathbf{r}\left(t\right)\right\Vert ^{2}}_{\text{First Term}}+
  2
  \underbrace{\left(\mathbf{r}\left(t+1\right)-\mathbf{r}\left(t\right)\right)^{\top}\mathbf{r}\left(t\right)}_{\text{Second Term}}
\end{equation}

The high-level approach is to bound the two terms of the above expansion for $\mathbf{r}(t)$ and then use a telescoping argument to bound \(\bfr(t)\) for all \(t > 0\). Below we provide the main arguments; for a complete proof of the second term's bound, please refer to Appendix \ref{appendix: proof_of_key_lemma}.

\subsection{Bounding the First Term}
Using \(\log(1+x) \leq x\) for all \(x>0\), we expand the first term as follows:
\begin{align*}
    \left\Vert \mathbf{r}\left(t+1\right)-\mathbf{r}\left(t\right)\right\Vert ^2
    & \leq\eta^{2}\left\Vert \nabla\mathcal{R}\left(\mathbf{w}\left(t\right)\right)\right\Vert ^{2}+\left\Vert \hat{\mathbf{w}}\right\Vert ^{2}t^{-2} + 2\eta\hat{\mathbf{w}}^{\top}\nabla\mathcal{R}\left(\mathbf{w}\left(t\right)\right)\log\left(1+t^{-1}\right) \\
    & \leq\eta^{2}\left\Vert \nabla\mathcal{R}\left(\mathbf{w}\left(t\right)\right)\right\Vert ^{2}+\left\Vert \hat{\mathbf{w}}\right\Vert ^{2}t^{-2}
\end{align*}
Obtaining the second inequality requires proving that
\begin{align} \label{eq: connective_tissue_inequality}
2\eta\hat{\mathbf{w}}^{\top}\nabla\mathcal{R}\left(\mathbf{w}\left(t\right)\right)\log\left(1+t^{-1}\right) \leq 0, \text{or equivalently, } \hat{\mathbf{w}}^{\top}\nabla\mathcal{R}\left(\mathbf{w}\left(t\right)\right) < 0
\end{align}
We will spend the rest of this subsection going over the complete proof of this inequality. 

First we state the following lemma (derived in Appendix \ref{appendix: proof_of_yutong}) that gives us a useful expression for the gradient of the risk w.r.t. 
$\mathbf{W}$:
\begin{lemma}\label{lemma: perm_risk_gradient}
For any \(\bfW \in \bbR^{d \times K}\), we have that
    \(
\nabla\mathcal{R}(\mathbf{W}) = \sum_{i=1}^{N}\mathbf{x}_{i}\nabla\psi\left(\mlc_{y_{i}}\mathbf{D}\mathbf{W}^{\top}\mathbf{x}_{i}\right)^{\top}\mlc_{y_{i}}\mathbf{D}
\).
\end{lemma}

This expression involves weight matrix $\mathbf{W}$. However the inequality we set out to prove (Eqn. \eqref{eq: connective_tissue_inequality}) is in terms of $\mathbf{w} = \vect(\mathbf{W})$. Throughout our main result proof, these two different forms -- weight matrix versus vectorization of that matrix -- will each be useful in different situations. Thus, to shuttle back and forth between these forms, the following well-known identity is useful:
\begin{lemma}
\label{lemma: general_vectorized_trace_identity}
For equally sized matrices $\bfM$ and $\bfN$, we have $\vect(\bfM)^{\top}\vect(\bfN) = \tr(\bfM^{\top}\bfN)$.
\end{lemma}

Now we can prove our inequality of interest, i.e., Eqn. \eqref{eq: connective_tissue_inequality}. 
\begin{lemma}
\label{lemma: connective_inequality}
(Multiclass generalization of \citet[Lemma 1]{soudry2018implicit}) For any PERM loss that is \(\beta\)-smooth, strictly decreasing, and non-negative, (Assumption \ref{assumption: loss_smoothness_dec_nonneg}) and Assumption \ref{assumption: loss_exp_tail}, and for almost all linearly separable datasets (Assumption \ref{assumption: dataset_linearly_separable}), we have $\hat{\mathbf{w}}^{\top}\nabla\mathcal{R}(\mathbf{w}(t)) < 0$.
\end{lemma}
\begin{proof}
Define matrix $\hat{\mathbf{W}}$ such that $\hat{\mathbf{w}} = \vect(\hat{\mathbf{W}})$. Since 
$\mathbf{w}(t) = \vect(\mathbf{W}(t))$, 
Lemma \ref{lemma: general_vectorized_trace_identity} 
implies
\begin{align}
\hat{\mathbf{w}}^{\top}\nabla\mathcal{R}(\mathbf{w}(t)) = \tr(\hat{\mathbf{W}}^{\top}\nabla\mathcal{R}(\mathbf{W}(t))) \label{eq: risk_gradient_trace_identity}
\end{align}

To see how the PERM framework allows for a simple generalization of binary results, we will compare our multiclass proof side-by-side with the binary proof discussed in \citet[Lemma 1]{soudry2018implicit}. In the binary case, we have
$\mathcal{R}\left(\mathbf{w}\right) = \sum^{N}_{i=1}\psi\left(y_i\mathbf{w}^{\top}\mathbf{x}_i\right) \implies \nabla\mathcal{R}\left(\mathbf{w}\right) =
\sum^{N}_{i=1}\psi'\left(y_i\mathbf{w}^{\top}\mathbf{x}_i\right)y_i\mathbf{x}_i$.
Thus
\(  \hat{\bfw}^{\top}\nabla\mathcal{R}\left(\mathbf{w}\right) =
\sum^{N}_{i=1}\psi'\left(y_i\mathbf{w}^{\top}\mathbf{x}_i\right)y_i\hat{\bfw}^{\top}\mathbf{x}_i.
\)
In the multiclass case, the analogous quantity is
$\tr(\hat{\mathbf{W}}^{\top}\nabla\mathcal{R}(\mathbf{W}(t)))$ which can be computed as
\[
  \sum_{i=1}^{N}  \tr(\hat{\mathbf{W}}^{\top}\mathbf{x}_{i}\nabla\psi\left(\mlc_{y_{i}}\mathbf{D}\mathbf{W}(t)^{\top}\mathbf{x}_{i}\right)^{\top}\mlc_{y_{i}}\mathbf{D})     =
 \sum_{i=1}^{N} \nabla\psi\left(\mlc_{y_{i}}\mathbf{D}\mathbf{W}(t)^{\top}\mathbf{x}_{i}\right)^{\top}
  \mlc_{y_{i}}\mathbf{D}\hat{\mathbf{W}}^{\top}\mathbf{x}_{i}.
\]
In the multiclass proof we used the risk gradient from Lemma \ref{lemma: perm_risk_gradient} as well as the cyclic property of the trace operator. Then we dropped the trace because \(\nabla\psi\left(\mlc_{y_{i}}\mathbf{D}\mathbf{W}(t)^{\top}\mathbf{x}_{i}\right)^{\top}\mlc_{y_{i}}\mathbf{D}\hat{\mathbf{W}}^{\top}\mathbf{x}_{i}\) is a scalar (since \(\nabla\psi\left(\cdot\right) \in \mathbb{R}^{K-1}, \mlc_{y_{i}}\mathbf{D}\mathbf{W}(t)^{\top}\mathbf{x}_{i} \in \mathbb{R}^{K-1}\)).
For illustrative purpose, we place the rest of the proof, in both the binary and multiclass setting, side-by-side:

\begin{minipage}{0.42\textwidth}
\textbf{Binary}:
$\hat{\mathbf{w}}^{\top}\nabla\mathcal{R}(\mathbf{w}(t))$.
    
Focusing on just the $i$-th term of this sum:
\begin{align*}
\psi'\left(y_i\mathbf{w}(t)^{\top}\mathbf{x}_i\right)y_i\hat{\mathbf{w}}^{\top}\mathbf{x}_i  
\end{align*}

$\psi$ is assumed to be strictly decreasing, i.e. $\psi'\left(y_i\mathbf{w}(t)^{\top}\mathbf{x}_i\right) < 0$. The dataset is linearly separable, so $y_i\hat{\mathbf{w}}^{\top}\mathbf{x}_i \geq 1$. Thus we obtain a sum (from $i = 1$ to $N$) of negative terms.

\end{minipage}%
\hfill
\begin{minipage}{0.5\textwidth}
\begin{tabular}{|p{0.91\textwidth}}
\textbf{Multiclass}: $\tr(\hat{\mathbf{W}}^{\top}\nabla\mathcal{R}(\mathbf{W}(t)))$.

Focusing on just the $i$-th term of this sum:
\begin{align*}
     \nabla\psi\left(\mlc_{y_{i}}\mathbf{D}\mathbf{W}(t)^{\top}\mathbf{x}_{i}\right)^{\top}
  \mlc_{y_{i}}\mathbf{D}\hat{\mathbf{W}}^{\top}\mathbf{x}_{i}
\end{align*}

$\psi$ is assumed to be strictly decreasing, i.e. $\nabla\psi\left(\mlc_{y_{i}}\mathbf{D}\mathbf{W}(t)^{\top}\mathbf{x}_{i}\right) \prec \mathbf{0}$. The dataset is linearly separable, so $\mlc_{y_{i}}\mathbf{D}\hat{\mathbf{W}}^{\top}\mathbf{x}_{i} \succeq 1$. Thus we obtain a sum (from $i = 1$ to $N$) of negative terms.
\end{tabular}
\end{minipage}

Thus we see how the PERM framework allows us to essentially mirror the binary proof. In Remark \ref{remark: perm_usefulness}, we elaborate more on the necessity of the relative margin form here.
\end{proof}

Lemma \ref{lemma: connective_inequality} directly implies the auxiliary inequality we set out to prove (see Eqn.~\eqref{eq: connective_tissue_inequality}). Thus we obtain:
\begin{align}
\left\Vert \mathbf{r}\left(t+1\right)-\mathbf{r}\left(t\right)\right\Vert ^{2} \leq \eta^{2}\left\Vert \nabla\mathcal{R}\left(\mathbf{w}\left(t\right)\right)\right\Vert ^{2}+\left\Vert \hat{\mathbf{w}}\right\Vert ^{2}t^{-2} \label{eq: r(t+1)_norm_first_term_bound}
\end{align}


\begin{remark}\label{remark: perm_usefulness}
Let us see what happens to our proof if we just used the general risk form in Eqn. \eqref{eq: general_risk} without the PERM framework. First, we need an expression for the gradient of the risk: \(\nabla\mathcal{R}\left(\mathbf{W}\right) = \sum^{N}_{i=1}\mathbf{x}_i\nabla\mathcal{R}_{y_i}\left(\mathbf{W}^{\top}\mathbf{x_i}\right)^{\top}\). Proceeding similarly to the binary case, we focus on just the $i$-th term of $\tr\left(\hat{\mathbf{W}}^{\top}\nabla\mathcal{R}\left(\mathbf{W}\right)\right)$:
\begin{align*}
\tr\left(\hat{\mathbf{W}}^{\top}\mathbf{x}_i\nabla\mathcal{R}_{y_i}\left(\mathbf{W}^{\top}\mathbf{x_i}\right)^{\top}\right)
 = \tr\left(\nabla\mathcal{R}_{y_i}\left(\mathbf{W}^{\top}\mathbf{x_i}\right)^{\top}\hat{\mathbf{W}}^{\top}\mathbf{x}_i\right) = \nabla\mathcal{R}_{y_i}\left(\mathbf{W}^{\top}\mathbf{x_i}\right)^{\top}\hat{\mathbf{W}}^{\top}\mathbf{x}_i
\end{align*}
From here it is not clear how to proceed. The linear separability condition (Assumption \ref{assumption: dataset_linearly_separable}) is not useful anymore- it does not make a statement about the scores in the vector $\hat{\mathbf{W}}^{\top}\mathbf{x}_i$, but rather their \emph{relative margins} (produced by the multiplication $\mlc_{y_{i}}\mathbf{D}\hat{\mathbf{W}}^{\top}\mathbf{x}_{i}$).
\end{remark}

\subsection{Bounding the Second Term}\label{section:bounding-the-second-term}


In the previous subsection we established a bound on the first term of Eqn. \eqref{eq: r(t+1) norm}. Here we sketch the main arguments required to bound the second term, i.e. $\left(\mathbf{r}\left(t+1\right)-\mathbf{r}\left(t\right)\right)^{\top}\mathbf{r}\left(t\right)$. For more details please refer to Appendix \ref{appendix: proof_of_key_lemma}.
We state our final bound below as a lemma:
\begin{lemma}
\label{lemma: r(t+1)_second_term_bound} (Generalization of \citet[Lemma 20]{soudry2018implicit}) Define $\theta$ to be the minimum SVM margin across all datapoints and classes, i.e. $\theta = \min_{k}\left[\min_{n\notin\mathcal{S}_k}\tilde{\mathbf{x}}_{n,k}^\top\hat{\mathbf{w}}\right]>1$. Then
\begin{equation}
\exists C_{1},C_{2},t_{1}:\,\forall t>t_{1}:\,\left(\mathbf{r}\left(t+1\right)-\mathbf{r}\left(t\right)\right)^{\top}\mathbf{r}\left(t\right)\leq C_{1}t^{-\theta} + C_{2}t^{-2}\, .
\label{eq: key_lemma_general_case}
\end{equation}
\end{lemma}
A remark is in order on the difference of the above result to \citet[Lemma 20]{soudry2018implicit}:
 on a high-level, we are able to generalize the argument of \citet[Lemma 20]{soudry2018implicit} to account for \emph{both binary and multiclass classification}, as well as general PERM ET losses beyond just CE.

We now proceed with the proof sketch. The first step is to rewrite \(\left(\mathbf{r}\left(t+1\right)-\mathbf{r}\left(t\right)\right)^{\top}\mathbf{r}\left(t\right)\) as
\begin{align}
& \left(-\eta\nabla\mathcal{R}\left(\mathbf{w}\left(t\right)\right)-\hat{\mathbf{w}}\left[\log\left(t+1\right)-\log\left(t\right)\right]\right)^{\top}\mathbf{r}\left(t\right)
\qquad \because
\mbox{Definition of 
\(\bfr(t)\) in \Cref{equation:residual}}
\nonumber \\
& =\hat{\mathbf{w}}^{\top}\mathbf{r}(t)\left(t^{-1}-\log\left(1+t^{-1}\right)\right) + \tr\left(\left(-\eta\sum_{i=1}^{N}\mathbf{x}_{i}\nabla\psi\left(\mlc_{y_{i}}\mathbf{D}\mathbf{W}(t)^{\top}\mathbf{x}_{i}\right)^{\top}\mlc_{y_{i}}\mathbf{D}\right)^{\top}\mathbf{R}(t)\right) \nonumber \\ & \qquad \qquad -  t^{-1}\hat{\mathbf{w}}^{\top}\mathbf{r}(t)
\qquad \because \mbox{Expression for \(\nabla\mathcal{R}\left(\mathbf{w}\left(t\right)\right)\) from \Cref{lemma: connective_inequality}
}
\label{eq: r(t+1)_norm_second_term_intermediate_form_main_text}
\end{align}
We defer the bound on the first term \(\hat{\mathbf{w}}^{\top}\mathbf{r}(t)\left(t^{-1}-\log\left(1+t^{-1}\right)\right)\) of \Cref{eq: r(t+1)_norm_second_term_intermediate_form_main_text} to the appendix, and instead focus on the second two terms. 
Using the cyclic property of the trace, the term in the above final line involving the trace can be further simplified  as:
\begin{equation}
  \label{equation:trace-inner-product-relative-margin-with-the-residual}
  \sum_{i=1}^{N}-\nabla\psi\left(\mlc_{y_{i}}\mathbf{D}
      \mathbf{W}(t)^{\top}\mathbf{x}_{i}\right)^{\top}
  \mlc_{y_{i}}\mathbf{D}\mathbf{R}(t)^{\top}\mathbf{x}_{i}
\end{equation}
Note that for each \(i \in [N]\), a summand in 
  \Cref{equation:trace-inner-product-relative-margin-with-the-residual}
is an inner product between two \((K-1)\)-dimensional vectors, i.e., \(-\nabla\psi\left(\mlc_{y_{i}}\mathbf{D}
      \mathbf{W}(t)^{\top}\mathbf{x}_{i}\right)\)
      and
      \(
  \mlc_{y_{i}}\mathbf{D}\mathbf{R}(t)^{\top}\mathbf{x}_{i}\).
To proceed, to expand this inner product out as 
\begin{equation}
  \label{equation:trace-inner-product-pseudo-index}
\sum^{N}_{i=1}\sum_{k \in [K] \backslash \{y_i\}}\pseudoindex{-\nabla\psi\left(\mlc_{y_{i}}\mathbf{D}\mathbf{W}(t)^{\top}\mathbf{x}_{i}\right)}_{k}
    \pseudoindex{\mlc_{y_{i}}\mathbf{D}\mathbf{R}(t)^{\top}\mathbf{x}_{i}}_{k}
\end{equation}

\begin{remark}
Here, \(\pseudoindex{ \cdot  }_{k} : \bbR^{K-1} \to \bbR\) is defined as the coordinate projection such that
  \(
    \pseudoindex{ \mlc_{y_{i}} \bfD \bfW^{\top} \bfx_{i} }_{k}
    =
    \tilde{\bfx}_{i,k}^{\top} \bfw
\). Note that \(\pseudoindex{ \cdot  }_{k}\) implicitly depends on \(i\) (the \(\tilde{\bfx}_{i,y_i}\) 0-entry is omitted). But we abuse notation for brevity. Please see Appendix \ref{appendix: pseudo-index} for a more precise definition.
\end{remark}
Using \Cref{eq: multiclass_SVM_coefficients} and \Cref{equation:w-tilde-defining-condition} we express the last two terms in \Cref{eq: r(t+1)_norm_second_term_intermediate_form_main_text} as
\begin{align}
&
 \Big(\sum^{N}_{i=1}
 \sum_{k \in [K] \setminus \{y_i\}}
  \pseudoindex{-\nabla\psi\left(\mlc_{y_{i}}\mathbf{D}\mathbf{W}(t)^{\top}\mathbf{x}_{i}\right)}_{k}
  \pseudoindex{\mlc_{y_{i}}\mathbf{D}\mathbf{R}(t)^{\top}\mathbf{x}_{i}}_{k}\Big) - t^{-1}\hat{\mathbf{w}}^{\top}\mathbf{r}(t) \nonumber
 \\
 &=
 \sum^{N}_{i=1}
 \sum_{k \in [K] \setminus \{y_i\}}
 \big(
 \pseudoindex{-\nabla\psi\left(\mlc_{y_{i}}\mathbf{D}\mathbf{W}(t)^{\top}\mathbf{x}_{i}\right)}_{k}  -
  t^{-1}\exp(-\tilde{\bfw}^\top \tilde{\bfx}_{i,k})  
 \mathbbm{1}_{\{i \in S_{k}\}} 
 \big)
 \tilde{\bfx}_{i,k}^\top \bfr(t)
 \label{equation:reduction-to-SHNGS}
\end{align}
Finally, to upper bound the above expression, we consider a single tuple \((i,k)\) case-by-case, depending on the sign of \(\tilde{\bfx}_{i,k}^\top \bfr(t)\). This is the step where the upper and lower bounds in \Cref{def: proposed_multiclass_exp_tail} come in.
\Cref{lemma:pseudo-label-exponential-tail-bounds} in the appendix essentially applies \Cref{def: proposed_multiclass_exp_tail} to the relative margins to yield
  \begin{align}
    &\pseudoindex{- \nabla \tp(\mlc_{y_i} \bfD \bfW(t)^{\top} \bfx_{i})}_{k}
    \le
    \exp(- \tilde{\bfx}_{i,k}^{\top} \bfw(t) ), \qquad \mbox{and}
    \label{equation:upper-bound-main-text}\\
    &\textstyle
    \pseudoindex{- \nabla \tp(\mlc_{y_i} \bfD \bfW(t)^{\top} \bfx_{i})}_{k}
    \ge
    (1 - \sum_{r \in [K] \setminus \{y_{i}\}} \exp(-\tilde{\bfx}_{i,r}^{\top} \bfw(t)))
    \exp(- \tilde{\bfx}_{i,k}^{\top} \bfw(t) )
    \label{equation:lower-bound-main-text}
  \end{align}
  for all \(k \in [K]\setminus \{y_i\}\).
We use
\Cref{def: proposed_multiclass_exp_tail}'s exponential tail bounds by proving
that the relative margins
\(\tilde{\bfx}_{i,k}^{\top} \bfw(t)\) that appear in
\Cref{lemma:pseudo-label-exponential-tail-bounds} eventually become positive. This is true due to the following lemma (see Appendix \ref{appendix: infinity_lemma_proof} for the proof, which again mirrors the binary case):
\begin{lemma}\label{lemma: relative_margin_infinity}
    (Multiclass generalization of \citet[Lemma 1]{soudry2018implicit}) Consider any linearly separable dataset, and any PERM loss with template $\psi$ that is convex, \(\beta\)-smooth, strictly decreasing, and non-negative. 
    For all $k \in \{1, ..., K\},$ let $\mathbf{w}_{k}(t)$ be the gradient descent iterates at iteration $t$ for the $k^{th}$ class. Then $\forall i \in \{1, ..., N\}, \forall j \in \{1, ..., K\} \backslash \{y_{i}\}: \lim_{t\rightarrow\infty}(\mathbf{w}_{y_{i}}(t)-\mathbf{w}_{j}(t))^{\top}\mathbf{x}_{i} \rightarrow \infty$.
\end{lemma}
This lemma lets us use the exponential tail bounds with any finite \(u_{\pm}\). To conclude, we apply 
the upper \eqref{equation:upper-bound-main-text}
and lower bounds 
\eqref{equation:lower-bound-main-text} to the summation in \Cref{equation:reduction-to-SHNGS}, and reduce the problem to that of \citet[Appendix E]{soudry2018implicit}, thereby proving \Cref{lemma: r(t+1)_second_term_bound}. See our \Cref{appendix: proof_of_key_lemma} for details.



\subsection{Tying It All Together}
We use the logic of  \citet[Appendix A.2]{soudry2018implicit} to conclude the analysis. Define 
\begin{align*}
 C=\sum_{t=0}^{\infty}\left\Vert \mathbf{r}\left(t+1\right)-\mathbf{r}\left(t\right)\right\Vert ^{2} 
\leq \sum_{t=0}^{\infty}\eta^{2}\left\Vert \nabla\mathcal{R}\left(\mathbf{w}\left(t\right)\right)\right\Vert ^{2}+\left\Vert \hat{\mathbf{w}}\right\Vert ^{2}t^{-2}.
\end{align*}
In the latter inequality we used Eqn. \eqref{eq: r(t+1)_norm_first_term_bound}. Thus, $C$ is bounded because from \citet[Lemma 10]{soudry2018implicit}, we know that 
$\sum_{t=0}^{\infty}\left\Vert \nabla\mathcal{R}\left(\mathbf{w}\left(t\right)\right)\right\Vert ^{2}<\infty$. Here we note that \citet[Lemma 10]{soudry2018implicit} requires the ERM objective \(\mathcal{R}\left(\bfw\right)\) to be \(\beta'\)-smooth for some positive \(\beta'\). It is easy to show that if the loss is \(\beta\)-smooth, then \(\mathcal{R}\left(\bfw\right)\) is \(\beta\sigma^{2}_{\text{max}}\left(\bfX\right)\)-smooth. This explains the learning rate condition \(\eta < 2/\left(\beta\sigma^{2}_{\text{max}}\left(\bfX\right)\right)\) in our theorem. Also, a $t^{-p}$ power series converges for any $p > 1$. 

Recalling the initial expansion of $\Vert \mathbf{r}(t+1) \Vert$ from Eqn.~\eqref{eq: r(t+1) norm}:
\begin{align}
\left\Vert \mathbf{r}\left(t+1\right)\right\Vert ^{2} & =\left\Vert \mathbf{r}\left(t+1\right)-\mathbf{r}\left(t\right)\right\Vert ^{2}+2\left(\mathbf{r}\left(t+1\right)-\mathbf{r}\left(t\right)\right)^{\top}\mathbf{r}\left(t\right)+\left\Vert \mathbf{r}\left(t\right)\right\Vert ^{2}.
\end{align}

Combining the bounds in Eqn.~\eqref{eq: r(t+1)_norm_first_term_bound} and Lemma \ref{lemma: r(t+1)_second_term_bound}
into Eqn.~\eqref{eq: r(t+1) norm}, we find 
\begin{align*}
\,  \left\Vert \mathbf{r}\left(t\right)\right\Vert ^{2}-\left\Vert \mathbf{r}\left(t_{1}\right)\right\Vert ^{2}
  =\sum_{u=t_{1}}^{t-1}\left[\left\Vert \mathbf{r}\left(u+1\right)\right\Vert ^{2}-\left\Vert \mathbf{r}\left(u\right)\right\Vert ^{2}\right]
 \leq C+2\sum_{u=t_{1}}^{t-1}\left[C_{1}u^{-\theta}+C_{2}u^{-2}\right].
\end{align*}
Therefore,
$\left\Vert \mathbf{r}\left(t\right)\right\Vert $ is bounded, which proves our main theorem.

\section{Limitations}
Here we describe some of our work's limitations/possible future research directions. We note that these questions have been analyzed for the binary classification setting, but not for multiclass.


 \paragraph{Non-ET losses} In our paper we only analyze multiclass implicit bias for losses with the ET property. Another possible line of future work is to analyze the gradient descent dynamics for non-ET losses. \citet{nacson2019convergence} and \citet{ji2020gradient} prove that in the binary setting, ET and well-behaved super-polynomial tailed losses ensure convergence to the maximum-margin direction, while other losses may converge to a different direction with poor margin. Is such a characterization possible in the multiclass setting? 

\paragraph{Other gradient-based methods}This paper only analyzes vanilla gradient descent. Another line of work involves exploring implicit bias effects of other gradient-based methods, such as those characterized in \citet{gunasekar2018characterizing}. \citet{nacson2022stochastic} uses similar proof techniques to prove results for SGD, which is prevalent in practice and often generalizes better than vanilla GD (\citep{amir2021sgd}). 

 \paragraph{Non-asymptotic analysis} Our result proves that the gradient descent predictors \emph{asymptotically} do not overfit. However, in the binary classification case, \citet{shamir2021gradient} goes one step further and proves that for gradient-based methods, throughout the entire training process (not just asymptotically), both the empirical risk and the generalization error decrease at an essentially optimal rate (or remain optimally constant). Does the same phenomenon occur in the multiclass setting?

\section{Conclusion}
We use the permutation equivariant and relative margin-based (PERM) loss framework to provide an multiclass extension of the binary ET property. On a high level, while the binary ET bounds the negative derivative of the loss, our multiclass ET bounds each negative partial derivative of the PERM template $\psi$. We demonstrate our definition's validity for multinomial logistic loss, multiclass exponential loss, and PairLogLoss. We develop new techniques for analyzing multiclass gradient descent, and apply these to generalize binary implicit bias results to the multiclass setting. Our main result is that for almost all linearly separable multiclass datasets and a suitable ET PERM loss, the gradient descent iterates directionally converge towards the hard-margin multiclass SVM solution.
 
Our proof techniques in this paper demonstrate the power of the PERM framework to facilitate extensions of known binary results to multiclass settings and provide a unified treatment of both binary and multiclass classification. Thus it is possible that the binary results discussed in the Limitations section can also be extended using the PERM loss framework. In the future we would like to consider more complex settings that have been analyzed primarily for the binary case, such as non-separable data (\citet{ji2019implicit}) and two-layer neural nets (\citet{lyu2021gradient}).

\begin{ack}
CS was supported in part by the National Science Foundation under award 2008074, and by the Department of Defense, Defense Threat Reduction Agency under award HDTRA1-20-2-0002. The research of DS was funded by the European Union (ERC, A-B-C-Deep, 101039436). Views and opinions expressed are however those of the author only and do not necessarily reflect those of the European Union or the European Research Council Executive Agency (ERCEA). Neither the European Union nor the granting authority can be held responsible for them. DS also acknowledges the support of the Schmidt Career Advancement Chair in AI. YW was supported in part by the Eric and Wendy Schmidt AI in Science Postdoctoral Fellowship, a Schmidt Futures program.
\end{ack}


\appendix

\section{Discussion of \citet{lyu2019gradient}}\label{appendix: discussion_of_lyu_li}
\cite{lyu2019gradient} allow the class score functions to be linear classifiers, e.g., \(\bfw_{k}^{\top} \bfx_{i}\), but also nonlinear, e.g.,
``cubed'' linear classifier \((\bfw_{k}^{\top} \bfx_{i})^{3}\). By shifting the cubing operation to the loss, we can view the implicit regularization result of \cite{lyu2019gradient} as a result for losses beyond the cross entropy. 
This resulting loss is rather exotic and we are not aware of it being used in the literature; it is interesting nevertheless. However, the optimization problem would become non-convex, so convergence would not necessarily be to a global minimum:
\[
\min_{\bfw} \frac{1}{2}\|\bfw\|^{2} \quad \mbox{s.t. \(\left(\bfw_{y_i}^{\top}\bfx_i\right)^{3}-\left(\bfw_{k}^{\top}\bfx_i\right)^{3} \geq 1\) for all \(i \in [N], j \in [K]\backslash\{y_i\}\)}
\]
Moreover, the decision region for the \(k\)-th class, i.e., the set of  \(\bfx \in \bbR^{d}\) such that \((\bfw_{k}^{\top} \bfx)^{3} > (\bfw_{j}^{\top} \bfx)^{3}\) for all \(j \ne k\), is an intersection of sets constructed via cubic hypersurfaces.

More precisely, the \(k\)-th decision region can be written as
\[
\{\bfx \in \bbR^{d} : (\bfw_k^\top \bfx)^3 = \mathrm{argmax}_{j \in [K]} (\bfw_j^\top \bfx)^3\}
=
\bigcap_{j \in [K] : j \ne k} 
\{ \bfx \in \bbR^{d} : (\bfw_{k}^{\top} \bfx)^{3} > (\bfw_{j}^{\top} \bfx)^{3}\}
\]

Let us define \(\mathcal{H}_j :=
\{ \bfx \in \bbR^{d} : (\bfw_{k}^{\top} \bfx)^{3} =(\bfw_{j}^{\top} \bfx)^{3}\}
\). Note that \(\mathcal{H}_j \subseteq \bbR^{d}\) is the zero set of degree \(3\) polynomials with variables in \(\bfx\), hence, a cubic hypersurface. Now, the set
\(\{ \bfx \in \bbR^{d} : (\bfw_{k}^{\top} \bfx)^{3} > (\bfw_{j}^{\top} \bfx)^{3}\}
\) is a subset of the set-theoretic complement of \(\mathcal{H}_j\) in \(\bbR^{d}\). Thus, the decision regions are complicated geometric objects, compared to the classical hard-margin SVM.

\section{Matrix Calculus}\label{appendix: matrix_calculus}
This section of the appendix establishes matrix identities that will be useful for us to calculate the gradient/Hessian of the empirical risk objective \(\risk(\bfw)\).

\myheader{Vector-input scalar-output function}
Suppose \(f: \bbR^{n} \to \bbR\) is a continuously differentiable function.
Let \(\bfx =
[
x_{1}, \cdots,  x_{n}
]^{\top} \in \mathbb{R}^{n}\) be a vector of variables for differentiation.
Define the (column) vector of partial derivatives w.r.t.\  \(\bfx\):
\(\partial_{\bfx} := \begin{bmatrix}
\frac{\partial}{\partial_{x_{i}}}
\end{bmatrix}_{i \in [n]}
\).
The \emph{gradient} of \(f\), denoted \(\nabla f\), is the function \begin{equation}
  \label{equation:definition:VISO-function-gradient}
\nabla f: \bbR^{n} \to \bbR^{n}, \quad \mbox{where} \quad
  \nabla f(\bfx) = \partial_{\bfx} f(\bfx)
=
\begin{bmatrix}
  \frac{\partial f}{\partial x_{1}} (\bfx) & \cdots &
\frac{\partial f}{\partial x_{n}} (\bfx)
\end{bmatrix}^{\top}.
\end{equation}

Suppose that \(f\) is twice continuously differentiable.
The \emph{Hessian} of \(f\), denoted \(\nabla^{2} f\), is the function \begin{equation}
  \label{equation:definition:VISO-function-hessian}
\nabla^{2} f: \bbR^{n} \to \bbR^{n\times n}, \quad \mbox{where} \quad
  \nabla^{2} f(\bfx)
=
\begin{bmatrix}
  \frac{\partial^{2} f}{\partial x_{i} \partial x_{j}} (\bfx)
\end{bmatrix}_{i,j \in [n]}.
\end{equation}



\myheader{Matrix-input scalar-output function}
Let \(f: \bbR^{m \times n} \to \bbR\) be a differentiable function.
Let \(\bfX = [x_{ij}]_{i\in [m], j \in [n]} \in \mathbb{R}^{m \times n}\) be an arbitrary matrix. Define the matrix of partial derivatives w.r.t.\  \(\bfX\):
\[
\partial_{\bfX} := \begin{bmatrix}
\frac{\partial}{\partial_{x_{ij}}}
\end{bmatrix}_{i \in [m], j \in [n]}
\]
Define the gradient of \(f\), denoted \(\nabla f\), to be the function
\begin{equation}
  \label{equation:definition:MISO-function-gradient}
  \nabla f: \bbR^{m\times n} \to \bbR^{m\times n}, \quad \mbox{where}
  \quad
  \nabla(f)(\bfX) := \partial_{\bfX} f(\bfX)
  =
  \begin{bmatrix}
\frac{\partial }{\partial_{x_{ij}}}f(\bfX)
\end{bmatrix}_{i \in [m], j \in [n]}.
\end{equation}
We do not  define the Hessian of a matrix-input scalar-output function \(f : \bbR^{m\times n} \to \bbR\). Instead, we will define the Hessian for its \emph{vectorization} \(\vect(f) :\bbR^{m n} \to \bbR\).

\begin{definition}[Vectorization operator]\label{definition:calculus:vectorization}
Let \(\vect\) denote the vectorization operator by stacking the columns of a vector. In other words,
if \(\bfA \in \mathbb{R}^{m \times n}\) is a matrix with columns \(\bfa_{1},\dots, \bfa_{n} \in \bbR^m\), then
\[
  \vect(\bfA) :=
  \begin{bmatrix}
    \bfa_{1}^{\top}
    &
    \bfa_{2}^{\top}
    & \cdots &
    \bfa_{n}^{\top}
  \end{bmatrix}^{\top}.
\]
\end{definition}

  \begin{definition}[Vectorization of a matrix-input function]
    Let \(f : \bbR^{m\times n} \to \bbR\) be a matrix-input function,
    we define \(\vect(f) : \bbR^{mn} \to \bbR\) to be the vector-input function such that
    \[
      f(\bfA) = \vect(f)(\vect(\bfA)).
    \]
In particular, if \(f\) is already a vector-input function, then \(\vect(f) = f\).
  \end{definition}
See
  \cite[Ch.5-\S 15]{magnus2019matrix}.
Below in \Cref{lemma:hessian-formula-generic}, we give a convenient formula to calculate the Hessian of \(\risk(\bfw)\), the vectorization of \(\risk(\bfW)\).



The following relates the vectorization operator with the Kronecker product:
\begin{lemma}\label{lemma:calculus:vectorization-and-kronecker-product-identity}
Let \(\bfA \in \bbR^{p \times n}, \bfB \in \bbR^{n\times m}, \bfC \in \bbR^{m\times q} \) be matrices.
Then
  \[
    \vect(\bfA \bfB \bfC) = (\bfC^{\top} \otimes \bfA) \vect(\bfB).
  \]
\end{lemma}
\begin{proof}
  This is \cite[Theorem 2.2]{magnus2019matrix}.
\end{proof}

\subsection{Special case of the chain rule for linear functions}

\begin{proposition}\label{proposition:calculus:gradient-chain-rule-linear}
Let \(\bfM \in \bbR^{m \times n}\) be a matrix.
Let \(f : \bbR^{m} \to \bbR\) be a continuously differentiable function and define
\(g : \bbR^{n} \to \bbR\)  by
\(  g(\bfx) := f(\bfM \bfx).
\)
Then
  \[
    \nabla g(\bfx) = \bfM^{\top}\nabla f(\bfM \bfx),
\quad 
\mbox{and}\quad
    \nabla^{2}g(\bfx) = \bfM^{\top}\nabla^{2} f(\bfM \bfx) \bfM.
  \]
\end{proposition}
\begin{proof}
See \cite[Ch.9-\S 13]{magnus2019matrix} for the first identity and
  \cite[Ch.10-\S 8]{magnus2019matrix} for the second identity.
\end{proof}

The next  two results will be referred to as the ``gradient formula'' and the ``Hessian formula'', respectively, for the function \(g(\bfX) := f(\bfA \bfX^{\top} \bfB)\).

\begin{lemma}
\label{lemma: Yutong_matrix_chain_rule}
Let \(f : \bbR^{p \times q} \to \bbR\) be a matrix-input scalar-output differentiable function with Jacobian denoted \(\nabla f : \bbR^{p \times q} \to \bbR^{p \times q}\). Let \(\bfA \in \bbR^{p \times n}\), \(\bfX \in \bbR^{m\times n}\), and \(\bfB \in \bbR^{m\times q}\).
Define a function \(g : \bbR^{m\times n} \to \bbR\)
by \(g(\bfX) := f(\bfA \bfX^{\top} \bfB)\).
Then
\[
\nabla g(\bfX)=
\partial_{\bfX} f(\bfA\bfX^{\top}\bfB) = \bfB\nabla f(\bfA\bfX^{\top}\bfB)^{\top}\bfA.
\]
\end{lemma}

\begin{lemma}\label{lemma:hessian-formula-generic}
  Let \(f: \bbR^{p} \to \bbR\) be a vector-input scalar-output twice differentiable function.
Let \(\bfA \in \bbR^{p \times n}, \bfX \in \bbR^{m\times n}\) be matrices and \(\bfb \in \bbR^{m} \) be a (column) vector.
  Let \(\bfV\) be another matrix with the same shape as \(\bfX\).
  Let \(\bfx := \vect(\bfX)\)
  and \(\bfv := \vect(\bfV)\).
  Define \(g(\bfX):= f(\bfA \bfX^{\top}\bfb)\) and let \(\overline{g} = \vect(g)\) be the vectorization of \(g\).
  Then we have the following formula for
  computing
  \(
    \bfv^{\top}\nabla^{2} \overline{g}( \bfx )
    \bfv
  \):
  \[
    \bfv^{\top}\nabla^{2} \overline{g}( \bfx )
    \bfv
    =
    (\bfA \bfV^{\top} \bfb)^{\top}
    \nabla^{2} f(\bfA \bfX^{\top} \bfb)
    \bfA \bfV^{\top} \bfb^{\top}.
  \]
  \end{lemma}

\subsection{Proof of the gradient formula: Lemma~\ref{lemma: perm_risk_gradient}}\label{appendix: proof_of_yutong}

In the notation of Section 2.8.1 of the Matrix Cookbook, define matrix  \(\mathbf{U} \in \mathbb{R}^{p\times q}\) by
  \(\mathbf{U} := \mathbf{A} \mathbf{X}^{\top} \mathbf{B}
  \). Note that \(\mathbf{U}\) is a function of \(\mathbf{X}\).
  Then by Eqn.~(137) of the Matrix Cookbook, we have for each \((i,j) \in [m] \times [n]\)
  \[
      \frac{\partial}{\partial X_{ij}}
f(\mathbf{A} \mathbf{X}^{\top} \mathbf{B}) =
      \frac{\partial}{\partial X_{ij}}
f(\mathbf{U}) =
\mathrm{Tr}
\left[
  \left(
  \frac{\partial f(\mathbf{U})}{\partial \mathbf{U}}
  \right)^{\top}
      \frac{\partial \mathbf{U}}{\partial X_{ij}}
\right]
  \]
  Note that by definition, we have
  \(
  \frac{\partial f(\mathbf{U})}{\partial \mathbf{U}}
  =
  \nabla f(\mathbf{U}).
  \) Therefore
\[
      \frac{\partial}{\partial X_{ij}}
f(\mathbf{A} \mathbf{X}^{\top} \mathbf{B}) =
\mathrm{Tr}
\left[
    \nabla f(\mathbf{U})^{\top}
      \frac{\partial \mathbf{U}}{\partial X_{ij}}
\right]
  \]

  Next, write \(\mathbf{U} =
  \begin{bmatrix}
    U_{k\ell}
  \end{bmatrix}_{k \in [p],\ell \in [q]}
  \) in the ``matrix-comprehension'' notation.
  Recall that \({U}_{k\ell}\), i.e., the \((k,\ell)\)-th entry of \(\bfU\), is precisely computed by 
\(
\mathbf{A}[k,:]( \mathbf{X}^{\top} \mathbf{B})[:,\ell]
=
\mathbf{A}[k,:] \mathbf{X}^{\top} \mathbf{B}[:,\ell]\).
  For each \(k,\ell \in [p] \times [q]\), we have
  \[
      \frac{\partial {U}_{k\ell}}{\partial X_{ij}}
      =
      \frac{\partial
        (
        \mathbf{A}[k,:] \mathbf{X}^{\top} \mathbf{B}[:,\ell]
        )
      }{\partial X_{ij}}
  \]
  where ``\([k,:]\)''
  and
  ``
  \([:,\ell]\)
  ''
  denote taking the \(k\)-th row vector and \(\ell\)-th column vector, respectively.
  Now, by Eqn.~(71) of the Matrix Cookbook, we have the following expression of the matrix-partial derivative as an outer product
  \[
      \frac{\partial {U}_{k\ell}}{\partial \mathbf{X}}
      =
      \frac{\partial
\mathbf{A}[k,:] \mathbf{X}^{\top} \mathbf{B}[:,\ell]
      }{\partial \mathbf{X}}
      =
      \mathbf{B}[:,\ell] \mathbf{A}[k,:].
  \]
  From this, it follows that computing the entry-wise partial derivative at \(X_{ij}\) is simply obtained by indexing at \((i,j)\), i.e.,
\(      \frac{\partial {U}_{k\ell}}{\partial X_{ij}}
      =
      \mathbf{B}[i,\ell] \mathbf{A}[k,j]
      =
      \mathbf{A}[k,j]
      \mathbf{B}[i,\ell] 
\) (we emphasize that this is just a product of two scalars).
  Thus,
\(      \frac{\partial \mathbf{U}}{\partial X_{ij}}
      =
       \mathbf{A}[:,j] \mathbf{B}[i,:]
\).
  Consequently,
\[
      \frac{\partial}{\partial X_{ij}}
f(\mathbf{A} \mathbf{X}^{\top} \mathbf{B}) =
\mathrm{Tr}
\left[
  \nabla f(\mathbf{U})^{\top}
       \mathbf{A}[:,j] \mathbf{B}[i,:]
\right]
=
\mathbf{B}[i,:]
  \nabla f(\mathbf{U})^{\top}
       \mathbf{A}[:,j].
  \]
  In other words,
\(
      \frac{\partial}{\partial \mathbf{X}}
f(\mathbf{A} \mathbf{X}^{\top} \mathbf{B}) 
=
\mathbf{B}
  \nabla f(\mathbf{U})^{\top}
       \mathbf{A}
  \)
.

For our purposes, we replace \(f\) with \(\psi\), \(\bfA\) with \(\mlc_{y_{i}}\bfD\), \(\bfX\) with \(\bfW\), and \(\bfB\) with \(\mathbf{x}_i\). Thus we obtain
\[
\nabla\mathcal{R}\left(\mathbf{W}\right) = \sum^{N}_{i=1}\frac{\partial}{\partial \bfW}\psi\left(\mlc_{y_{i}}\bfD\bfW^{\top}\mathbf{x}_i\right) = \sum^{N}_{i=1}\mathbf{x}_i\nabla\psi\left(\mlc_{y_{i}}\bfD\bfW^{\top}\mathbf{x}_i\right)^{\top}\mlc_{y_{i}}\bfD
\]
as desired. \hfill \(\square\)

\subsection{Proof of Hessian formula: Lemma~\ref{lemma:hessian-formula-generic}}

Our goal is to calculate the Hessian of \(\vect(g)\). First, we note that by definition \[
\vect(g)(\bfx) =
    \vect(g)(\vect(\bfX)) =
    \vect(f)
    (
     \vect(\bfA \bfX^{\top} \bfb)
    )
  \]
  Note that the last equality 
  is simply
  \( \vect(f)
    (
     \vect(\bfA \bfX^{\top} \bfb)
    )
    =
    f(
     \bfA \bfX^{\top} \bfb)\), but we work in the more general case of a matrix \(\bfB\) right now.
We will need to simplify \(\vect(\bfA \bfX^{\top} \bfb)\).
It is more convenient during the first phase of the proof viewing \(\bfb\) as a \(m \times 1\) matrix and denote it using uppercase letter \(\bfB\). 
  First, applying \Cref{lemma:calculus:vectorization-and-kronecker-product-identity} to \(\vect(\bfA \bfX^{\top} \bfB)\), we get
  \[
    \vect(\bfA \bfX^{\top} \bfB) = (\bfB^{\top} \otimes \bfA) \vect(\bfX^{\top})
  \]

  However, \(\vect(\bfX^{\top})
  \ne \vect(\bfX)\) in general.
However, these two expressions are related using the \emph{commutation matrix}:

  \begin{definition}[Commutation matrix]
    \label{definition:calculus:commutation-matrix}
    Define \(\bfK_{m,n}\) to be the permutation matrix in \(\bbR^{mn\times mn}\) such that
    \(\bfK_{m,n} \vect(\bfA) = \vect(\bfA^{\top})\) for all matrices \(\bfA \in \bbR^{m\times n}\).
  \end{definition}
  See \cite[Ch.3-\S 7]{magnus2019matrix}.
Below, we drop the subscripts in \Cref{definition:calculus:commutation-matrix} and simply write \(\bfK := \bfK_{m,n}\).
Now, we have
  \[
    \vect(\bfA \bfX^{\top} \bfB) = (\bfB^{\top} \otimes \bfA) \bfK^{\top}\vect(\bfX)
    =
    (\bfB^{\top} \otimes \bfA) \bfK^{\top} \bfx
  \]

Thus
\[
\vect(g)(\bfx) =
    \vect(g)(\vect(\bfX)) =
    \vect(f)
    (
    (\bfB^{\top} \otimes \bfA) \bfK^{\top} \bfx
    )
  \]

  By \Cref{proposition:calculus:gradient-chain-rule-linear}, we have
\begin{align*}
&\nabla^{2} \vect(g) (\bfx)\\
&=
(
(\bfB^{\top} \otimes \bfA) \bfK^{\top}
)^{\top}
\nabla^{2} \vect(f)
    (
    (\bfB^{\top} \otimes \bfA) \bfK^{\top}
    \bfx
    )
(\bfB^{\top} \otimes \bfA) \bfK^{\top}
\\
&=
(
(\bfB^{\top} \otimes \bfA) \bfK^{\top}
)^{\top}
\nabla^{2} \vect(f)
    (
  \vect(\bfA \bfX^{\top} \bfB)
  )
(\bfB^{\top} \otimes \bfA) \bfK^{\top}.
\end{align*}

From this, we see that (recall that \(\bfv = \vect(\bfV)\))
\[
\bfv^\top \nabla^{2} \vect(g) (\bfx) \bfv
=
(
(\bfB^{\top} \otimes \bfA) \bfK^{\top}\bfv
)^{\top}
\nabla^{2} \vect(f)
    (
  \vect(\bfA \bfX^{\top} \bfB)
  )
(\bfB^{\top} \otimes \bfA) \bfK^{\top}\bfv.
\]

Now, by \Cref{lemma:calculus:vectorization-and-kronecker-product-identity}, we have
\[(\bfB^{\top} \otimes \bfA) \bfK^{\top}\bfv
=
(\bfB^{\top} \otimes \bfA) \bfK^{\top}\vect(\bfV)
=
\vect(
\bfA \bfV^\top \bfB
).
\]
Now, since \(\bfB = \bfb\) is just a vector, we have
\[
\vect(
\bfA \bfV^\top \bfB
)
=
\bfA \bfV^\top \bfb.
\]
and
\[
\nabla^{2} \vect(f)
    (
  \vect(\bfA \bfX^{\top} \bfB)
  )
  =
  \nabla^{2} f
    (
  \bfA \bfX^{\top} \bfb
  )
\]
Putting it all together, we get the desired equality. \hfill \(\square\)

\section{PERM Losses That Satisfy Assumptions \ref{assumption: loss_smoothness_dec_nonneg} and \ref{assumption: loss_exp_tail}}\label{appendix: losses_that_satisfy_assumptions}

\subsection{Cross-Entropy}
By \citet[Example 1]{wang2023unified},
the cross-entropy loss $\mathcal{L}_{y}\left(\mathbf{v}\right) =-\log\left(\frac{\exp\left(v_y\right)}{\sum_{k=1}^{K}\exp\left(v_k\right)}\right)$ has template $\psi\left(\mathbf{u}\right) = \log\left(1+\sum_{k=1}^{K-1}\exp\left(-u_{k}\right)\right)$.
We calculate the partial derivatives:
\begin{align}
& \frac{\partial \psi}{\partial{u}_{i}}\left(\mathbf{u}\right) \nonumber \\
&= 
-\frac{\exp\left(-u_{i}\right)}{1 + \sum_{k=1}^{K-1}\exp\left(-u_{k}\right)}\label{eq: CE_template_gradient_form1}
\\&=
-\frac{1}{1 + (C_{i}+1)\exp\left(u_{i}\right)} \quad \mbox{where \(C_{i} = \sum_{k \in [K-1]:k \neq i}\exp\left(-u_{k}\right).\)} \label{eq: CE_template_gradient_form2}
\end{align}

\subsubsection{Convexity}\label{appendix: CE_convexity}
Let us analyze the entries of the Hessian of the template, i.e. \(\nabla^{2} \psi\left(\mathbf{u}\right)\). Let \([\mathbf{A}]_{l,m}\) denote the element of \(\mathbf{A}\) at the \(l\)-th row and \(m\)-th column. We get for all \(i, j \in [K-1]\) where \(j \neq i\):
\begin{align*}
    & \left[\nabla^{2} \psi\left(\mathbf{u}\right)\right]_{i,i} = \frac{\partial^{2} \psi\left(\mathbf{u}\right)}{\partial u_{i}^{2}} = \frac{\left(C_{i}+1\right)e^{-u_i}}{\left(1 + \sum^{K-1}_{k=1}e^{-u_k}\right)^{2}} \\
    & \left[\nabla^{2} \psi\left(\mathbf{u}\right)\right]_{i,j} = \left[\nabla^{2} \psi\left(\mathbf{u}\right)\right]_{j,i} = \frac{\partial^{2} \psi\left(\mathbf{u}\right)}{\partial u_{i}u_{j}} = \frac{-e^{-u_i-u_j}}{\left(1 + \sum^{K-1}_{k=1}e^{-u_k}\right)^{2}}
\end{align*}

From the definition of \(C_{i}\), this implies that:
\begin{align*}
    \left[\nabla^{2} \psi\left(\mathbf{u}\right)\right]_{i,i} = \sum_{j \in [K-1], j \neq i} \left|\left[\nabla^{2} \psi\left(\mathbf{u}\right)\right]_{i,j}\right| + \frac{e^{-u_i}}{\left(1 + \sum^{K-1}_{k=1}e^{-u_k}\right)^{2}}
\end{align*}

Thus, the Hessian is a symmetric \emph{diagonally dominant} matrix, and hence is positive semi-definite.

\subsubsection{\(\beta\)-smoothness}\label{appendix: CE_beta_smoothness}
For any diagonally dominant matrix \(\mathbf{B}\), let \(\left|\mathbf{B}\right|\) be the matrix obtained by taking the absolute value of each element of \(\mathbf{B}
\), that is:
\[
\left[\left|\mathbf{B}\right|\right]_{l,m}=\left|\left[\mathbf{B}\right]_{l,m}\right| \quad \mbox{for all \(l,m\)}.
\]
Additionally, let \(\mathsf{diag}(\cdot) : \bbR^{p} \to \bbR^{p \times p}\) (for any \(p \in \mathbb{N}\)) be the function that maps a vector to a diagonal matrix in the obvious way.

Then we have the following lemma:
\begin{lemma}\label{lemma: diagonally_dominant}
Let \(\mathbf{B}' := \mathsf{diag}\left(\left|\mathbf{B}\right|\mathbf{1}\right)\) where \(\mathbf{1}\) is the appropriately-sized vector of all-1's. Then \(\mathbf{B} \preceq \mathbf{B'}\).
\end{lemma}
\begin{proof}
This can be proven simply by observing that \(\mathbf{B'}-\mathbf{B}\) is symmetric and diagonally dominant (and thus positive semi-definite).
\end{proof}

Lemma \ref{lemma: diagonally_dominant} can be directly applied to analyze the Hessian (and eventually bound its maximum eigenvalue). Define \(\mathbf{H'} = \mathsf{diag}\left(\left|\nabla^{2} \psi\left(\mathbf{u}\right)\right|\mathbf{1}\right)\). In other words, from Eqn. \eqref{eq: CE_template_gradient_form2}:
\begin{align}
    &\left[\mathbf{H'}\right]_{i,i} = \sum^{K-1}_{k=1}\left|\nabla^{2} \psi\left(\mathbf{u}\right)\right|_{i,k} = \frac{\left(2C_{i}+1\right)e^{-u_i}}{\left(1+\sum^{K-1}_{k=1}e^{-u_k}\right)^2} \label{eq: CE_diag_dom_diagonal_entry} \\
    &\left[\mathbf{H'}\right]_{i,j} = 0 \nonumber
\end{align}

Thus, by directly applying Lemma \ref{lemma: diagonally_dominant}, we obtain
\begin{align*}
     \nabla^{2} \psi\left(\mathbf{u}\right) \preceq \mathbf{H'}
\end{align*}

So now since \(\mathbf{H'}\) is defined to be a diagonal matrix, all that's left to do is bound the diagonal entries by a positive constant. First note that from the definition of \(C_{i}\), it follows that \(1+\sum^{K-1}_{k=1}e^{-u_k} = C_{i}+1+e^{-u_i}\). Combining this with Eqn. \eqref{eq: CE_diag_dom_diagonal_entry}, we get:
\[
\frac{\left(2C_{i}+1\right)e^{-u_i}}{\left(\left(C_{i}+1\right)+e^{-u_i}\right)^2} = \frac
{\left(2C_{i}+1\right)}{\left(\left(C_{i}+1\right)+e^{-u_i}\right)\left(\left(C_{i}+1\right)e^{u_i}+1\right)}
\]
We can find a global minimum of the denominator of the above expression and thus arrive at an upper bound for the expression. Differentiating with respect to \(u_i\) and setting to 0 yields a single critical point at \(u_i = -\log\left(C_{i}+1\right)\), which produces a value of \(4\left(C_{i}+1\right)\) when substituted in the denominator (this is a global minimum of the denominator expression).
Thus, we get \[
\left[\mathbf{H'}\right]_{i,i} \leq \frac{\left(2C_{i}+1\right)}{4\left(C_{i}+1\right)} = \frac{1}{2} - \frac{1}{4\left(C_{i}+1\right)}
\]
In the binary i.e. \(K=2\) case, \(C_{i}=0\), so our bound is exactly \(1/4\). However, in the multiclass case (i.e. \(K>2\)), \(C_{i}\) can be arbitrarily large. Setting \(C_{i} = \infty\) yields a final upper bound of \(1/2\).

So our final bound can be summarized as follows: \[\|\nabla^{2} \psi\left(\mathbf{u}\right)\|_{2} \leq \left[\mathbf{H}'\right]_{i,i} \leq \begin{cases}
1/4 & ,\,\mathrm{if}\,K=2\\
1/2 & ,\,\mathrm{if}\,K > 2.
\end{cases}\]
Thus, \(\beta=1/4\) for binary cross-entropy (logistic loss), but \(\beta=1/2\) for K-class cross-entropy.

\subsubsection{Exponential Tail}\label{appendix: CE_exp_tail}
We claim that for the cross-entropy Definition~\ref{def: proposed_multiclass_exp_tail} holds
with \(u_\pm = 0\) and \(c = a = 1\). We are interested in analyzing the (negative) gradient of the template. From Eqn. \ref{eq: CE_template_gradient_form1}:
\begin{align*}
    &-\frac{\partial \psi}{\partial{u}_{i}}\left(\mathbf{u}\right) = \frac{e^{-u_i}}{1 + \sum^{K-1}_{k=1}e^{-u_k}}\\ 
    &\leq e^{-u_i}\\
    &\geq e^{-u_i}\left(1-\sum^{K-1}_{k=1}e^{-u_k}\right) \quad \because \mbox{\(\forall x \geq 0, \frac{1}{1+x} \geq 1-x\)}
\end{align*}

This proves that the cross-entropy loss satisfies Definition~\ref{def: proposed_multiclass_exp_tail} with \(u_\pm = 0\) and \(c=1\).

\subsection{Multiclass Exponential Loss \citep{mukherjee2013theory}}
The multiclass exponential loss $\mathcal{L} : \mathbb{R}^{K}  \rightarrow \mathbb{R}$ can be written as \(
\mathcal{L}_y(\mathbf{v}) = \sum_{k \in \left[K\right]: k \neq y} \exp(-({v}_{y} - {v}_k))\).
Thus, the template function $\psi: \mathbb{R}^{K-1} \rightarrow \mathbb{R}$ can be expressed as
\(
\psi(\mathbf{u}) = \sum_{i \in \left[K-1\right]} e^{-u_i}
\). The partial derivatives of the template are then simply:
\begin{align}
  \frac{\partial}{\partial u_{i}}\psi(\mathbf{u}) = -e^{-u_i} \quad \mbox{for all \(i \in [K-1]\)}. \label{eq: exp_loss_template_gradient}
\end{align}

\subsubsection{Convexity}\label{appendix: exp_loss_convexity}
We have
\begin{align}
    \nabla^2 \psi(\bfu) = \mathsf{diag}(\exp(-u_i ): i = 1,\dots, K-1).\label{eq: exp_loss_hessian}
\end{align}
The Hessian is a diagonal matrix with all diagonal entries positive. Hence it is positive definite.

\subsubsection{\(\beta\)-``smoothness''}\label{appendix: exp_loss_pseudo_smoothness}
Recall the identity derived in Lemma \ref{lemma:hessian-formula-generic}:
\[
    \bfv^{\top}\nabla^{2} \overline{g}( \bfx )
    \bfv
    =
    (\bfA \bfV^{\top} \bfb)^{\top}
    \nabla^{2} f(\bfA \bfX^{\top} \bfb)
    \bfA \bfV^{\top} \bfb^{\top}.
  \]

We are interested in the special case where \(\bfX \gets \bfW\) is a linear classifier (represented as a matrix) and \(\bfx \gets \vect(\bfW) = \bfw\) is its vectorization as in \Cref{section:main-result}. Moreover, \(g(\bfX)\) represents the risk \(\risk(\bfW)\), viewed as a \emph{matrix-input scalar-output} function (defined in Appendix \ref{appendix: matrix_calculus}), while \(\overline{g}(\bfx)\) represents the \emph{vectorized} risk \(\risk(\bfw)\), viewed as a \emph{vector-input scalar-output} function (defined in Appendix \ref{appendix: matrix_calculus}).
We will use the formula in \Cref{lemma:hessian-formula-generic} to calculate \(\bfv^\top \nabla^2 \risk(\bfw) \bfv\), where we substitute in
\[
  \bfA \gets \mlc_{y_{i}} \bfD \in \bbR^{(K-1)\times K}
  ,\qquad \bfb \gets \bfx_{i} \in \bbR^{d}, \qquad f \gets \psi.
\]
where \((\bfx_i,y_i)\) is a training sample and \(\psi\) is the template of a PERM loss. Since \(\nabla^2\) is linear (i.e., distributive over additions), we have by \Cref{lemma:hessian-formula-generic} that
\begin{align*}
&\bfv^\top \nabla^2 \risk (\bfw)
\bfv
=
\vect(\bfV)^\top \nabla^2 \risk (\bfw)
\vect(\bfV) \\
&=
\sum_{i=1}^{N}
(\mlc_{y_i}\bfD^\top \bfV \bfx_i)^\top
\nabla^2 \psi ( 
\mlc_{y_i}\bfD^\top \bfW \bfx_i
)
\mlc_{y_i}\bfD^\top \bfV \bfx_i.
\end{align*}
Note that \(\| \bfv\| = \| \vect(\bfV)\| = \| \bfV\|_F\) by the definitions of the Frobenius norm and vectorization. Thus, 
\begin{equation}
    \label{equation:risk-max-eigenvalue}
\max_{\bfv \in \mathbb{R}^{dK}: \| \bfv\|=1}
\bfv^\top \nabla^2 \risk (\bfw)
\bfv
=
\max_{\bfV \in \mathbb{R}^{d\times K}: \| \bfV\|_F=1}
\vect(\bfV)^\top \nabla^2 \risk (\bfw)
\vect(\bfV).
\end{equation}
We note that we never defined the Hessian of a \emph{matrix}-input function, i.e.,
we do not work with \(\nabla^2 \risk (\bfW)\).
Combining the two previous identities, we have proven
\begin{corollary}\label{corollary:hessian-of-risk-max-eigenvalue}
Let \(\risk(\bfW)\)  be the risk viewed as a matrix-input scalar-output function defined in \Cref{eq: risk_perm_form}. Let \(\risk(\bfw)\) be the vectorization of \(\risk(\bfW)\). Then we have
\begin{align*}
&\max_{\bfv \in \mathbb{R}^{dK}: \| \bfv\|=1}
\bfv^\top \nabla^2 \risk (\bfw)
\bfv \\
&=
\max_{\bfV \in \mathbb{R}^{d \times K}: \| \bfV\|_F=1}
\sum_{i=1}^{N}
(\mlc_{y_i}\bfD^\top \bfV \bfx_i)^\top
\nabla^2 \psi ( 
\mlc_{y_i}\bfD^\top \bfW \bfx_i
)
\mlc_{y_i}\bfD^\top \bfV \bfx_i.
\end{align*}

\end{corollary}

We have
\begin{align}
    \nabla^2 \psi(\bfu) = \mathsf{diag}(\exp(-u_i ): i = 1,\dots, K-1).
\end{align}
Let \(\mathsf{vdiag}(\cdot)\) be the ``inverse'' of 
\(\mathsf{diag}(\cdot)\), i.e., \(\mathsf{vdiag}(\cdot)\)  takes a diagonal matrix and returns the vector of the diagonal elements.

The max eigenvalue of the Hessian  of \(\risk(\bfw)\), i.e., 
\Cref{equation:risk-max-eigenvalue}, is computed below:
\begin{align*}
&\max_{\bfv \in \mathbb{R}^{dK}: \| \bfv \|  = 1}
\bfv^\top \nabla^2 \risk(\bfw) \bfv\\
&\overset{1}{=}
\max_{\bfV \in \mathbb{R}^{d \times K}: \| \bfV \|_F  = 1}
\sum_{i=1}^N
(\mlc_{y_i} \bfD \bfV^\top \bfx_i)^\top
\nabla^2 \psi(\mlc_{y_i} \bfD \bfW^\top \bfx_i)
\mlc_{y_i} \bfD \bfV^\top \bfx_i \quad \because \mbox{
\Cref{corollary:hessian-of-risk-max-eigenvalue}}
\\
&\overset{2}{=}
\max_{\bfV \in \mathbb{R}^{d \times K}: \| \bfV \|_F  = 1}
\sum_{i=1}^N
\tr\Big(
\nabla^2 \psi(\mlc_{y_i} \bfD \bfW^\top \bfx_i)
\underbrace{
\mlc_{y_i} \bfD \bfV^\top \bfx_i
(\mlc_{y_i} \bfD \bfV^\top \bfx_i)^\top
}_{(K-1)\mbox{-by-}(K-1)\mbox{ outer product}}
\Big)
\\
&
\overset{3}{=}
\max_{\bfV \in \mathbb{R}^{d \times K}: \| \bfV \|_F  = 1}
\sum_{i=1}^N
\underbrace{
\mathsf{vdiag}(
\nabla^2 \psi(\mlc_{y_i} \bfD \bfW^\top \bfx_i)
)^\top}_{\text{Vector of diagonal elements of } \nabla^2 \psi}
\underbrace{
(\mlc_{y_i} \bfD \bfV^\top \bfx_i)^{\odot2}
}_{\text{entrywise square}}
\\
&
\le
\max_{\bfV \in \mathbb{R}^{d \times K}: \| \bfV \|_F  = 1}
\sum_{i=1}^N
\risk(\bfW)
\mathbf{1}^\top
(\mlc_{y_i} \bfD \bfV^\top \bfx_i)^{\odot2}
\quad \because \mbox{replace each entry of \(\mathsf{vdiag}\)  with risk}
\end{align*}
In equality 2 we took trace of a scalar (the expression in equality 1 is a scalar, so taking the trace of it will not change the value) and used the cyclic property. For equality 3: as per \Cref{eq: exp_loss_hessian}, \(\nabla^{2}\psi\) is a diagonal matrix. Finally, in the last inequality, we bound each element of the diagonal vector (i.e. \(\exp\left(-u_i\right)\) for all \(i \in [K-1]\)).
Dropping the \(
\risk(\bfW)
\)
and
``\(\max_{\bfV \in \mathbb{R}^{d \times K}: \| \bfV \|_F  = 1}\)''
from the front:
\begin{align*}    
&
\sum_{i=1}^N \mathbf{1}^\top
(\mlc_{y_i} \bfD \bfV^\top \bfx_i)^{\odot2}\\
&= \sum_{i=1}^N
(\mlc_{y_i} \bfD \bfV^\top \bfx_i )^\top
\mlc_{y_i} \bfD \bfV^\top \bfx_i = \sum_{i=1}^N
\tr\left((\mlc_{y_i} \bfD \bfV^\top \bfx_i )^\top
\mlc_{y_i} \bfD \bfV^\top \bfx_i\right) \\
&=
  \sum_{i=1}^N
\tr
\Big(
\bfD^\top \mlc_{y_i}^\top
\mlc_{y_i} \bfD\bfV^\top \bfx_i \bfx_i^\top
\bfV
\Big) \quad \mbox{Note that  \(\mathbf{V}^{\top}\mathbf{x}_i \in \mathbb{R}^{K}\)}\\
&\le\sum_{i=1}^N
\|\bfD^\top \mlc_{y_i}^\top
\mlc_{y_i} \bfD
\|_F
\|
\bfV^\top \bfx_i
\bfx_i^\top
\bfV
\|_F
\quad \because \mbox{Cauchy-Schwarz inequality}\\
  \end{align*}
Note that we applied Cauchy-Schwarz to the inner product space \(\bbR^{K \times K}\) with inner product \(\langle \bfA, \bfB\rangle := \tr(\bfA^{\top} \bfB)\).
Now, continuing with the calculation:
  \begin{align*}
&\sum_{i=1}^N
\|\bfD^\top \mlc_{y_i}^\top
\mlc_{y_i} \bfD
\|_F
\|
\bfV^\top \bfx_i
\bfx_i^\top
\bfV
\|_F
\\
&\le\sum_{i=1}^N \|\mlc_{y_i} \bfD\|^{2}_{F}
\|\mathbf{V}^{\top}\|_{F}\|\mathbf{x}_i\mathbf{x}_i^{\top}\|_{F}\|\mathbf{V}\|_{F} \quad \because \mbox{\(\|\mathbf{A}\mathbf{B}\|_{F} \leq \|\mathbf{A}\|_{F}\|\mathbf{B}\|_{F}\)}\\
&\leq \left(2K-2\right)\sum_{i=1}^N \|\mathbf{x}_i\|^{2} \quad \because \mbox{Lemma \ref{lemma:frobenius-norm-of-mlc-D}, \(\|\mathbf{x}_i\mathbf{x}_i^{\top}\|_{F} = \|\mathbf{x}_i\|^{2}\), \(\|\mathbf{V}\|_{F}=1\)} \\
&\leq \left(2K-2\right)\left(\sum_{i=1}^N \|\mathbf{x}_i\|\right)^{2} \quad \because \mbox{for all \(a_j > 0\), \(M \geq 1\), \(\sum^{M}_{i=1} a^{2}_j \leq \left(\sum^{M}_{i=1} a_j\right)^{2}\)}
\end{align*}

Therefore we have proven that \(\|\nabla^{2}\mathcal{R}\left(\mathbf{w}\right)\|_{2} \leq B^2\mathcal{R}\left(\mathbf{w}\right)\), where 
\begin{align}
  B = \sqrt{\left(2K-2\right)} \sum_{i=1}^{N}\|\bfx_i\|.  \label{eq: exp_loss_B_defn}
\end{align}

Now we will also analyze the Euclidean norm of the gradient, for reasons that will become clear later.

\begin{align*}
    &\|\nabla\mathcal{R}\left(\bfw\right)\| = \|\nabla\mathcal{R}\left(\bfW\right)\|_{F}\\
    &=\left\Vert\sum^{N}_{i=1}\bfx_i\nabla\psi\left(\mlc_{y_i}\bfD\bfW^{\top}\bfx_i\right)^{\top}\mlc_{y_i}\bfD\right\Vert_{F}\\
    & \leq \sum^{N}_{i=1} \left\Vert\bfx_i\nabla\psi\left(\mlc_{y_i}\bfD\bfW^{\top}\bfx_i\right)^{\top}\mlc_{y_i}\bfD\right\Vert_{F} \quad \because \mbox{triangle inequality}\\
    & \leq \sum^{N}_{i=1} \|\bfx_i\|_{2}\left\Vert\nabla\psi\left(\mlc_{y_i}\bfD\bfW^{\top}\bfx_i\right)\right\Vert_{2}\|\mlc_{y_i}\bfD\|_{F} \quad \because \mbox{\(\|\bfA \bfB\|_{F} \leq \|\bfA\|_{F} \|\bfB\|_{F}\)}\\
    & = \sqrt{\left(2K-2\right)}\sum^{N}_{i=1} \|\bfx_i\|_{2}\left\Vert\nabla\psi\left(\mlc_{y_i}\bfD\bfW^{\top}\bfx_i\right)\right\Vert_{2} \quad \because \mbox{Lemma \ref{lemma:frobenius-norm-of-mlc-D}}\\
    & \leq \sqrt{\left(2K-2\right)} \sum^{N}_{i=1}\left\Vert\bfx_i\right\Vert\mathcal{R}\left(\bfw\right)  \quad \because \mbox{for all \(a_j > 0\), \(M \geq 1\), \(\sqrt{\sum^{M}_{i=1} a^{2}_j} \leq \sum^{M}_{i=1} a_j\)}
\end{align*}

Gradient descent is a special case of steepest descent with the Euclidean norm \citep{boyd2004convex}. Thus, we can apply \citet[Lemmas 11 \& 12]{gunasekar2018characterizing} to see that \(\nabla\mathcal{R}\left(\mathbf{w}\right) \rightarrow 0 \) even for multiclass exponential loss. Elaborating on this: these lemmas from \citet{gunasekar2018characterizing} assume a convex risk objective (which we have in the case of multiclass exponential loss). Additionally, they assume that \(\left\Vert\nabla\mathcal{R}\left(\bfw\right)\right\Vert \leq B\mathcal{R}\left(\bfw\right)\) and \( \left\Vert\nabla^{2}\mathcal{R}\left(\bfw\right)\right\Vert_{2} \leq B^{2}\mathcal{R}\left(\bfw\right) \). In the above section we prove these exact results with \(B\) as defined in Eqn. \ref{eq: exp_loss_B_defn}. Finally, \Cref{lemma:frobenius-norm-of-mlc-D} below proves that \(\|\mlc_{k} \bfD\|_{F} = \sqrt{2K-2}\) for all \(k \in [K-1]\).

In conclusion, by \citet[Lemma 11]{gunasekar2018characterizing}, if our learning rate \(\eta < \frac{1}{B^{2}\mathcal{R}\left(\bfw(0)\right)}\), we can use \citet[Lemma 10]{soudry2018implicit}.

Finally, we calculate \(\|\mlc_{k} \bfD\|_{F} \) which was used in several places above:

\begin{lemma}\label{lemma:frobenius-norm-of-mlc-D}
    For each \(k \in [K]\), we have
    \(\|\mlc_{k} \bfD\|_{F} = \sqrt{2(K-1)} \).
\end{lemma}
\begin{proof}
First, if \(k = K\), then \(\mlc_K\) is the identity matrix.
In this case, we have \(\mlc_{k} \bfD = \bfD=  [-\mathbf{I}_{K-1} \quad \mathbf{1}_{K-1}] \) is the negative \((K-1)\)-by-\((K-1)\) identity matrix concatenated with the all-ones vector. Thus, 
\[
\|\mlc_{k} \bfD\|_{F}^2
=
\| \bfD \|_F^2
=
\| \mathbf{I}_{K-1}\|_F^2 + 
\|  \mathbf{1}_{K-1}\|^2
=2(K-1)
\]
If \(k \ne K\), then 
    \[
    \mlc_{k} \bfD 
    =[-\mlc_{k} \quad \mlc_{k}\mathbf{1}_{K-1}]
    \]
    and so 
    \[
\|\mlc_{k} \bfD\|_{F}^2
=
\|\mlc_{k}\|_F^2 + 
\|  \mlc_{k}\mathbf{1}_{K-1}\|^2
\]
Now, we recall from the definition of \(\mlc_{k}\) (Definition 2.4 of \cite{wang2023unified}) that
\(\mlc_{k}\) is obtained by replacing the \(k\)-th column of the identity matrix by the ``all-negative-ones'' vector.
Thus 
\[
\|\mlc_{k}\|_F^2
=2(K-1) -1, \quad \mbox{and} \quad
\|\mlc_{k}\mathbf{1}_{K-1} \|_F^2
=\|-\bfe_k \|_F^2
=1.
\]
This proves \Cref{lemma:frobenius-norm-of-mlc-D}, as desired.
\end{proof}

\subsubsection{Exponential Tail}
From Eqn. \eqref{eq: exp_loss_template_gradient}, the negative partial derivative of the template is clearly always positive:
\[
-\frac{\partial}{\partial u_{i}}\psi(\mathbf{u}) = e^{-u_i} \geq 0.
\]
From the above, it is clear that the upper and lower bounds in Definition~\ref{def: proposed_multiclass_exp_tail} hold when \(u_{\pm} = 0\) and \(c=1\).

\subsection{PairLogLoss \citep{wang2022rank4class}}

Recall that the template of the PairLogLoss is \(\psi\left(\mathbf{u}\right) = \sum^{K-1}_{k=1}\log\left(1+\exp\left(-u_k\right)\right)\). By elementary calculus, we see that 
\begin{align*}
    \frac{\partial\psi(u)}{\partial u_k} = -\frac{e^{-u_k}}{1+ e^{-u_k}}.
\end{align*}
\subsubsection{Convexity}
We have
\begin{align*}
    \nabla^2 \psi(\bfu) = \mathsf{diag}\left(e^{u_i}/\left(1+e^{u_i}\right)^2: i = 1,\dots, K-1\right).
\end{align*}
The Hessian is a diagonal matrix with all diagonal entries positive. Hence it is positive definite.
\subsubsection{\(\beta\)-smoothness}
Notice that the partial derivative of the template is exactly the same expression as the derivative of the logistic loss (i.e. binary cross-entropy). Thus, the exact same proof as logistic loss can be used to prove \(\beta\)-smoothness for the PairLogLoss as well. Thus, from Appendix \ref{appendix: CE_beta_smoothness}, \(\beta = 1/4\) (logistic loss is simply the \(K=2\) case for cross-entropy).
\subsubsection{Exponential Tail}
\begin{align*}
    -\frac{\partial\psi(u)}{\partial u_k} = \frac{e^{-u_k}}{1+ e^{-u_k}} \leq e^{-u_k}
\end{align*}
This gives us the desired upper tail. As for the lower tail:
\begin{align*}
    &-\frac{\partial\psi(u)}{\partial u_k} = \frac{e^{-u_k}}{1+ e^{-u_k}} \\
    &\geq e^{-u_k}\left(1-e^{-u_k}\right) \quad \because \mbox{\(\frac{1}{1+x}\geq 1-x\) for all \(x \geq 0\)} \\
    & \geq e^{-u_k}\left(1-\sum^{K-1}_{i=1}e^{-u_i}\right)
\end{align*}
Thus, PairLogLoss satisfies Definition \ref{def: proposed_multiclass_exp_tail} with \(u_{\pm} = 0\) and \(c = a = 1\).

\section{Pseudo-index}\label{appendix: pseudo-index}

Note that $\bm{\Upsilon}_{y_{i}}\mathbf{D}\mathbf{R}(t)^{\top}\mathbf{x}_{i}$ produces a $(K-1)$-dimensional vector with entries of the form of $\left(\mathbf{r}_{y_{i}}(t) - \mathbf{r}_{k}(t)\right)^{\top}\mathbf{x}_{i}, \forall k \in [K]\backslash \{y_{i}\}$.
For $k \in [K]\backslash \{y_{i}\}$, let us represent the corresponding entry of the vector as $\pseudoindex{\bm{\Upsilon}_{y_{i}}\mathbf{D}\mathbf{R}(t)^{\top}\mathbf{x}_{i}}_{k}$.
Note that this indexing is not the same as the $k^{th}$ entry of the vectors, since the ${y_i}^{th}$ entry $\left(\mathbf{r}_{y_{i}}(t) - \mathbf{r}_{y_{i}}(t)\right)^{\top}\mathbf{x}_{i}$ is not present in the vector.
Similarly, let us define $\pseudoindex{-\nabla\psi\left(\bm{\Upsilon}_{y_{i}}\mathbf{D}\mathbf{W}(t)^{\top}\mathbf{x}_{i}\right)}_{k}$ to be the corresponding entry of $-\nabla\psi$.

This section makes this indexing trick rigorous.
\begin{lemma}\label{lemma:pseudo-index-existence}
  Let \(\bfW \in \bbR^{d \times K}\) be arbitrary  and \(\bfw := \vect(\bfW)\) be its vectorization.
  Let \((\bfx_{i}, y_{i})\) be a training sample.
  Then there exists a bijection that depends only on \(y_{i}\) that maps the entries of
  \[  \mlc_{y} \bfD \bfW^{\top} \bfx_{i} \in \bbR^{K-1}\]
to the elements of the set of ``\(y_{i}\)-versus-\(k\)'' relative margins, i.e.,
\(\{ \tilde{\bfx}_{i,k}^{\top} \bfw  \in \bbR : k \in [K] \setminus \{y_{i}\}\}
\).
\end{lemma}
The following definition makes the bijection from 
\Cref{lemma:pseudo-index-existence} concrete.
\begin{definition}[Pseudo-index]\label{definition:pseudo-index}
In the situation of \Cref{lemma:pseudo-index-existence},
define \(\pseudoindex{ \cdot  }_{i,k} : \bbR^{K-1} \to \bbR\) to be the coordinate projection such that
  \(
    \pseudoindex{ \mlc_{y_{i}} \bfD \bfW \bfx_{i} }_{i,k}
    =
    \tilde{\bfx}_{i,k}^{\top} \bfw
\).
  In other words,
\(\pseudoindex{ \cdot  }_{i,k}\) selects the \(y_{i}\)-versus-\(k\) relative margin.
When the sample index \(i\) is clear from context, we drop \(i\) from the subscript and simply write
\(\pseudoindex{ \cdot  }_{k}\).
\end{definition}

The pseudo-index is useful for working with the exponential tail bounds:
\begin{lemma}\label{lemma:pseudo-label-exponential-tail-bounds}
  In the situation of \Cref{lemma:pseudo-index-existence}, consider
  \(\nabla \tp (\mlc_{y_i} \bfD \bfW^{\top} \bfx_{i})\) which is the \((K-1)\)-dimensional vector of partial derivatives
  of the template evaluated at
  \(\mlc_{y_i} \bfD \bfW^{\top} \bfx_{i}\).
  If \(\tp\) satisfies
  \Cref{def: proposed_multiclass_exp_tail}, then
  \[
    \pseudoindex{- \nabla \tp(\mlc_{y_i} \bfD \bfW^{\top} \bfx_{i})}_{k}
    \le
    \exp(- \tilde{\bfx}_{i,k}^{\top} \bfw )
  \]
  and
  \[
    \textstyle
    \pseudoindex{- \nabla \tp(\mlc_{y_i} \bfD \bfW^{\top} \bfx_{i})}_{k}
    \ge
    (1 - \sum_{r \in [K] \setminus \{y_{i}\}} \exp(-\tilde{\bfx}_{i,r}^{\top} \bfw))
    \exp(-\tilde{\bfx}_{i,k}^{\top} \bfw )
  \]
  for all \(k \in [K]\setminus \{y_i\}\).

\end{lemma}

\subsection{Proofs of \Cref{lemma:pseudo-index-existence} and \Cref{lemma:pseudo-label-exponential-tail-bounds}}
In both lemmas, we work with a fixed sample, i.e., the index \(i\)  does not change. As such, we simply drop the index and write \(y\gets y_i\),
\(\bfx \gets  \bfx_i\), 
\(\tilde{\bfx}_k \gets  \tilde{\bfx}_{i,k}\),
and
\(\pseudoindex{ \cdot  }_{k} \gets \pseudoindex{ \cdot  }_{i,k}\).

Below, we fix \(k \in [K-1]\) throughout the proof.
Let \(\bfv := \bfw^{\top} \bfx = [v_{1} ,\, \dots, \, v_{K}]^{\top}\).
Note that
\begin{equation}
  \label{equation:margin-vector}
  \mlc_{y} \bfD^{\top} \bfW^{\top} \bfx
  =
  \mlc_{y}
  \begin{bmatrix}
   v_{K} - v_{1} \\ \vdots \\  v_{K} - v_{K-1}
  \end{bmatrix}
\end{equation}

We prove \Cref{lemma:pseudo-index-existence} by considering the case \(y=K\) and \(y \ne K\) separately.
First, let us consider the case when \(y=K\).
Then \(\mlc_{y}\) is the identity matrix and so
\[
  \mbox{
\(k\)-th component of
  \Cref{equation:margin-vector} is
  }
  =
  v_K - v_k
  =
  (
  \bfw_K - \bfw_k
  )^\top \bfx
  = \tilde{\bfx}_k^\top \bfw.
\]
Thus, we've proven  \Cref{lemma:pseudo-index-existence}  when \(y=K\).
In this case, \(\pseudoindex{ \cdot  }_{k}\) simply picks out the \(k\)-th entry of the input (\(K-1)\)-dimensional) vector.
In other words,
\begin{equation}
    \label{definition:pseudo-label-easy-case}
    \pseudoindex{\bfz}_k = z_k, \quad 
\mbox{for all \(\bfz = [z_1,\, \dots, z_{K-1}]^\top \in \bbR^{K-1}\) when \(y = K\).}
\end{equation}

By definition, we note that the \(k\)-th row of \(\mlc_{y}\) is
\[
\mlc_{y}[k,:]
=
\begin{cases}
  \bfe_{k} - \bfe_{y} &: k \ne y, \\
  - \bfe_{y} &: \mbox{otherwise}.
\end{cases}
\]
Thus, when \(y \ne K\)
\begin{align*}
  \mbox{
\(k\)-th component of
  \Cref{equation:margin-vector}
  }
  &=
\begin{cases}
  (v_K-v_n)  &: k \ne y, \\
  - (v_K-v_y)
           &: k = y.
\end{cases}\\
&=
\begin{cases}
  (\bfw_y- \bfw_k)^\top \bfx = \tilde{\bfx}_k &: k \ne y, \\
 (\bfw_y- \bfw_K)^\top \bfx = \tilde{\bfx}_K
           &: k = y.
\end{cases}
\end{align*}

Thus, we've proven  \Cref{lemma:pseudo-index-existence}  when \(y\ne K\) as well. In this case, \(\pseudoindex{ \cdot  }_{k}\) picks out the \(k\)-th element of the input (\(K-1)\)-dimensional) vector when \(k \ne y\). Otherwise when \(k = y\),  we have that \(\pseudoindex{ \cdot  }_{k}\) picks out the \(y\)-th element. More explicitly,
\begin{equation}
    \label{definition:pseudo-label-quote-hard-unquote-case}
    \pseudoindex{\bfz}_k = 
    \begin{cases}
        z_k &: k \ne y\\
        z_y &: k = y
    \end{cases}
    , \quad 
\mbox{for all \(\bfz = [z_1,\, \dots, z_{K-1}]^\top \in \bbR^{K-1}\)  when \(y \ne K\).}
\end{equation}

Next, we prove \Cref{lemma:pseudo-label-exponential-tail-bounds} by considering the case \(y=K\) and \(y \ne K\) separately. First, assume that we are in the \(y =K\) case.
From    \Cref{definition:pseudo-label-easy-case}, we get that
\[
 \pseudoindex{- \nabla \tp(\mlc_{y_i} \bfD^{\top} \bfW^{\top} \bfx_{i})}_{k}
 =
 -\frac{\partial \tp}{\partial u_k}(\mlc_{y_i} \bfD^{\top} \bfW^{\top} \bfx_{i})
\]

Now, 
we have that for \(\bfu = [u_1,\, \dots, \, u_{K-1}]^\top \in \bbR^{K-1}\), the upper and lower exponential tail bounds are 
\[
-\frac{\partial \psi}{\partial {u}_{k}}\left(\mathbf{u}\right) \leq c\exp(-u_{k}), \quad \mbox{and} \quad -\frac{\partial \psi}{\partial {u}_{k}}\left(\mathbf{u}\right) \geq c\left(1-\sum_{r \in [K-1]}\exp\left(-u_{r}\right)\right)\exp(-u_{k}).
\]
Letting \(\bfu :=\mlc_{y_i} \bfD^{\top} \bfW^{\top} \bfx_{i} \), using \Cref{lemma:pseudo-index-existence} and     \Cref{definition:pseudo-label-easy-case}, we immediately prove \Cref{lemma:pseudo-label-exponential-tail-bounds} in the case when \(y=K\). This is because we have \[\exp(-u_{k}) 
=
\exp(-\pseudoindex{\mlc_{y_i} \bfD^{\top} \bfW^{\top} \bfx_{i} }_k) 
= \exp(-\tilde{\bfx}_{i,k}^\top \bfw) \] and 
\[\sum_{r \in [K-1]}\exp\left(-u_{r}\right)
=
\sum_{r \in [K]\setminus \{y_i\}}\exp\left(-u_{r}\right)
=
\sum_{k \in [K]\setminus \{y_i\}}\exp\left(-\tilde{\bfx}_{i,k}^\top \bfw\right)
\]

When \(y=K\), a similar argument proves \Cref{lemma:pseudo-label-exponential-tail-bounds} using \Cref{definition:pseudo-label-easy-case} for the pseudo-index \(\pseudoindex{\cdot}_k\).

\section{Proof of Lemma \ref{lemma: relative_margin_infinity}}\label{appendix: infinity_lemma_proof}
Re-stating the lemma:

\textbf{Lemma \ref{lemma: relative_margin_infinity}.}
\textit{
    (Multiclass generalization of \citet[Lemma 1]{soudry2018implicit}) Consider any linearly separable dataset, and any PERM loss with template $\psi$ that is convex, \(\beta\)-smooth, strictly decreasing, and non-negative. 
    For all $k \in \{1, ..., K\},$ let $\mathbf{w}_{k}(t)$ be the gradient descent iterates at iteration $t$ for the $k^{th}$ class. Then $\forall i \in \{1, ..., N\}, \forall j \in \{1, ..., K\} \backslash \{y_{i}\}: \lim_{t\rightarrow\infty}(\mathbf{w}_{y_{i}}(t)-\mathbf{w}_{j}(t))^{\top}\mathbf{x}_{i} \rightarrow \infty$.
}

\begin{proof}
We know that $\lim_{t\rightarrow \infty}\nabla\mathcal{R}\left(\mathbf{w}\left(t\right)\right) = \mathbf{0}$ by \citet[Lemma 10]{soudry2018implicit}.

This implies that $\hat{\mathbf{w}}^{\top}\nabla\mathcal{R}(\mathbf{w}(t)) \rightarrow \mathbf{0}$. Following the same steps as in the proof of Lemma \ref{lemma: connective_inequality}, this is equivalent to saying:
\[
\hat{\mathbf{w}}^{\top}\nabla\mathcal{R}(\mathbf{w}(t)) = \tr(\nabla\psi\left(\bm{\Upsilon}_{y_{i}}\mathbf{D}\mathbf{W}(t)^{\top}\mathbf{x}_{i}\right)^{\top}\bm{\Upsilon}_{y_{i}}\mathbf{D}\hat{\mathbf{W}}^{\top}\mathbf{x}_{i}) \rightarrow 0.
\]
However, for linearly separable data we know that $\bm{\Upsilon}_{y_{i}}\mathbf{D}\hat{\mathbf{W}}^{\top}\mathbf{x}_{i} \succeq \mathbf{1}$ (since $\hat{\mathbf{W}}$ here is the hard-margin SVM solution). Thus for the above limit to be true, the limit
\[
\lim_{t\rightarrow \infty}\nabla\psi\left(\bm{\Upsilon}_{y_{i}}\mathbf{D}\mathbf{W}(t)^{\top}\mathbf{x}_{i}\right) = \mathbf{0}
\]
must hold.
By \Cref{proposition:filling-the-gap}, we have
\[
\lim_{t \rightarrow \infty} \bm{\Upsilon}_{y_{i}}\mathbf{D}\mathbf{W}(t)^{\top}\mathbf{x}_{i} = \boldsymbol{\infty} \qquad \forall i \in [N]
\]
where $\boldsymbol{\infty}$ is the  ``vector'' whose entries are all equal to infinity. 
This is equivalent to
\[
\lim_{t\rightarrow\infty}(\mathbf{w}_{y_{i}}(t)-\mathbf{w}_{j}(t))^{\top}\mathbf{x}_{i} = \infty \qquad \forall i \in [N], \forall j \in [K] \backslash \{y_i\}
\]
(since $\pseudoindex{\bm{\Upsilon}_{y_{i}}\mathbf{D}\mathbf{W}(t)^{\top}\mathbf{x}_{i}}_j = (\mathbf{w}_{y_{i}}(t)-\mathbf{w}_{j}(t))^{\top}\mathbf{x}_{i}$).
\end{proof}
Note that in the binary case, the above ``convergence-to-infinity'' condition is for a scalar quanity, where the assumption that the loss  be strictly decreasing and non-negative suffices. In the multiclass setting, we must ensure that all entries of the vector 
\(\bm{\Upsilon}_{y_{i}}\mathbf{D}\mathbf{W}(t)^{\top}\mathbf{x}_{i} \)
converges to infinity. This is a nontrivial result and is addressed by our \Cref{proposition:filling-the-gap}.
\section{Proof of Lemma \ref{lemma: r(t+1)_second_term_bound}}\label{appendix: proof_of_key_lemma}

Let us first re-state the lemma we want to prove. 

\textbf{Lemma \ref{lemma: r(t+1)_second_term_bound}}. \textit{(Generalization of \citet[Lemma 20]{soudry2018implicit}) Define $\theta$ to be the minimum SVM margin across all data points and classes, i.e., $\theta = \min_{k}\left[\min_{n\notin\mathcal{S}_k}\tilde{\mathbf{x}}_{n,k}^\top\hat{\mathbf{w}}\right]>1$. Then:
\begin{equation}
\exists C_{1},C_{2},t_{1}:\,\forall t>t_{1}:\,\left(\mathbf{r}\left(t+1\right)-\mathbf{r}\left(t\right)\right)^{\top}\mathbf{r}\left(t\right)\leq C_{1}t^{-\theta}+C_{2}t^{-2}\, \nonumber
\end{equation}
}
\begin{proof}
Proceeding the same way as \cite{soudry2018implicit}, we have
\begin{align}
& \left(\mathbf{r}\left(t+1\right)-\mathbf{r}\left(t\right)\right)^{\top}\mathbf{r}\left(t\right)\nonumber \\
& =\left(-\eta\nabla\mathcal{R}\left(\mathbf{w}\left(t\right)\right)-\hat{\mathbf{w}}\left[\log\left(t+1\right)-\log\left(t\right)\right]\right)^{\top}\mathbf{r}\left(t\right)\nonumber \\
& = \left(-\eta\nabla\mathcal{R}\left(\mathbf{w}\left(t\right)\right)\right)^{\top}\mathbf{r}(t) - \hat{\mathbf{w}}^{\top}\mathbf{r}(t)\log\left(1+t^{-1}\right)\nonumber \\
& =\hat{\mathbf{w}}^{\top}\mathbf{r}(t)\left(t^{-1}-\log\left(1+t^{-1}\right)\right) + \tr\left(\left(-\eta\sum_{i=1}^{N}\mathbf{x}_{i}\nabla\psi\left(\mlc_{y_{i}}\mathbf{D}\mathbf{W}(t)^{\top}\mathbf{x}_{i}\right)^{\top}\mlc_{y_{i}}\mathbf{D}\right)^{\top}\mathbf{R}(t)\right) \nonumber \\ & \qquad -  t^{-1}\hat{\mathbf{w}}^{\top}\mathbf{r}(t) \label{eq: r(t+1)_norm_second_term_intermediate_form}
\end{align}

The last equality is a new step required for our multiclass generalization, in which we used Lemma \ref{lemma: perm_risk_gradient} and introduced the matrices $\mathbf{W}(t)$ and $\mathbf{R}(t)$, where $\vect(\mathbf{W}(t)) = \mathbf{w}(t)$ and $\vect(\mathbf{R}(t))=\mathbf{r}(t)$. Let us focus just on the second term of this expansion.
\begin{align}
    & \tr\left(\left(-\eta\sum_{i=1}^{N} \mathbf{x}_{i}\nabla\psi\left(\mlc_{y_{i}}\mathbf{D}\mathbf{W}(t)^{\top}\mathbf{x}_{i}\right)^{\top}\mlc_{y_{i}}\mathbf{D}\right)^{\top}\mathbf{R}(t)\right) \nonumber\\
    & \overset{(1)}{=} \eta \tr\left(\sum_{i=1}^{N}\mathbf{R}(t)^{\top}\mathbf{x}_{i}\left(-\nabla\psi\left(\mlc_{y_{i}}\mathbf{D}\mathbf{W}(t)^{\top}\mathbf{x}_{i}\right)\right)^{\top}\mlc_{y_{i}}\mathbf{D}\right) \nonumber \\
    & \overset{(2)}{=} \eta \tr\left(\sum_{i=1}^{N}\left(-\nabla\psi\left(\mlc_{y_{i}}\mathbf{D}\mathbf{W}(t)^{\top}\mathbf{x}_{i}\right)\right)^{\top}\mlc_{y_{i}}\mathbf{D}\mathbf{R}(t)^{\top}\mathbf{x}_{i}\right) \label{eq: last_risk_trace_before_dot_product}
\end{align}

In step (1) we used the fact that for any square matrix $\mathbf{M}$, $\tr(\mathbf{M}) = \tr\left(\mathbf{M}^{\top}\right)$.  In step (2) we used the cyclic property of the trace. 


Similar to in the proof of Lemma \ref{lemma: connective_inequality}, the trace's cyclic property has enabled us to convert a matrix-product into a simple dot product. Since dot products are scalars, we can now drop the trace and rewrite our expression in Eqn. $\eqref{eq: last_risk_trace_before_dot_product}$ as a dot product: 
\begin{align*}
  & \eta \sum^{N}_{i=1}\sum_{k \in [K] \backslash \{y_i\}}\pseudoindex{-\nabla\psi\left(\mlc_{y_{i}}\mathbf{D}\mathbf{W}(t)^{\top}\mathbf{x}_{i}\right)}_{k}
    \pseudoindex{\mlc_{y_{i}}\mathbf{D}\mathbf{R}(t)^{\top}\mathbf{x}_{i}}_{k}
\end{align*}
Using this form, we can rewrite Eqn.~\eqref{eq: r(t+1)_norm_second_term_intermediate_form}:
\begin{align}
 \left(\mathbf{r}\left(t+1\right)-\mathbf{r}\left(t\right)\right)^{\top}\mathbf{r}\left(t\right) &= \hat{\mathbf{w}}^{\top}\mathbf{r}(t)\left(t^{-1}-\log\left(1+t^{-1}\right)\right) \nonumber \\
&\qquad + \eta \sum^{N}_{i=1}\sum_{k \in [K]\setminus \{y_i\}}
\pseudoindex{-\nabla\psi\left(\mlc_{y_{i}}\mathbf{D}\mathbf{W}(t)^{\top}\mathbf{x}_{i}\right)}_{k}
\pseudoindex{\mlc_{y_{i}}\mathbf{D}\mathbf{R}(t)^{\top}\mathbf{x}_{i}}_{k}
  \nonumber
  \\
  &\qquad-  t^{-1}\hat{\mathbf{w}}^{\top}\mathbf{r}(t)
\label{eq: r(t+1)_norm_second_term_final_form}
\end{align}
The first term \(\hat{\mathbf{w}}^{\top}\mathbf{r}(t)\left(t^{-1} - \log\left(t^{-1}\right)\right)\) is bounded in \cite[Eqn.~(139)]{soudry2018implicit}. We will focus on the second and third terms.
Recall by
\Cref{eq: multiclass_SVM_coefficients}
and
\Cref{equation:w-tilde-defining-condition}
that
\[
\hat{\bfw} =
\sum_{i=1}^N \sum_{k\in [K] \setminus \{y_i\}}
\alpha_{i,k} \mathbbm{1}_{\{i \in S_{k}\}}  \tilde{\bfx}_{i,k}
=
\sum_{i=1}^N \sum_{k\in [K] \setminus \{y_i\}}
\eta 
\exp(-\tilde{\bfw}^\top \tilde{\bfx}_{i,k}) \mathbbm{1}_{\{i \in S_{k}\}}  \tilde{\bfx}_{i,k}
\]
Thus, the third term on the RHS of \Cref{eq: r(t+1)_norm_second_term_final_form}
can be written as
\[
t^{-1}
\hat{\mathbf{w}}^{\top}\mathbf{r}(t)
=
\eta
\sum_{i=1}^N \sum_{k\in [K] \setminus \{y_i\}}
t^{-1}
\exp(-\tilde{\bfw}^\top \tilde{\bfx}_{i,k})  \tilde{\bfx}_{i,k}^\top \bfr(t)
\mathbbm{1}_{\{i \in S_{k}\}} 
\]
Therefore, the last two terms on the RHS of 
\Cref{eq: r(t+1)_norm_second_term_final_form}
can be written as
\begin{align*}
&
 \eta \sum^{N}_{i=1}
 \sum_{k \in [K] \setminus \{y_i\}}
  \pseudoindex{-\nabla\psi\left(\mlc_{y_{i}}\mathbf{D}\mathbf{W}(t)^{\top}\mathbf{x}_{i}\right)}_{k}
  \pseudoindex{\mlc_{y_{i}}\mathbf{D}\mathbf{R}(t)^{\top}\mathbf{x}_{i}}_{k}\nonumber -  t^{-1}\hat{\mathbf{w}}^{\top}\mathbf{r}(t)
 \\
 &=
 \eta \Big(
 \sum^{N}_{i=1}
 \sum_{k \in [K] \setminus \{y_i\}}
   \pseudoindex{-\nabla\psi\left(\mlc_{y_{i}}\mathbf{D}\mathbf{W}(t)^{\top}\mathbf{x}_{i}\right)}_{k}
   \pseudoindex{\mlc_{y_{i}}\mathbf{D}\mathbf{R}(t)^{\top}\mathbf{x}_{i}}_{k}\\
 & \qquad \qquad \qquad  \qquad \qquad -  t^{-1}\exp(-\tilde{\bfw}^\top \tilde{\bfx}_{i,k})  \tilde{\bfx}_{i,k}^\top \bfr(t)
 \mathbbm{1}_{\{i \in S_{k}\}} 
 \Big)
\end{align*}

Since \(\eta >0\) is constant, we ignore it below and consider only the term inside the parenthesis:

\begin{align}
&
 \sum^{N}_{i=1}
 \sum_{k \in [K] \setminus \{y_i\}}
  \pseudoindex{-\nabla\psi\left(\mlc_{y_{i}}\mathbf{D}\mathbf{W}(t)^{\top}\mathbf{x}_{i}\right)}_{k}
  \pseudoindex{\mlc_{y_{i}}\mathbf{D}\mathbf{R}(t)^{\top}\mathbf{x}_{i}}_{k} \nonumber \\
 & \qquad \qquad \qquad  \qquad \qquad -  t^{-1}\exp(-\tilde{\bfw}^\top \tilde{\bfx}_{i,k})  \tilde{\bfx}_{i,k}^\top \bfr(t)
 \mathbbm{1}_{\{i \in S_{k}\}} \nonumber 
 \\
 &\overset{(1)}{=}
 \sum^{N}_{i=1}
 \sum_{k \in [K] \setminus \{y_i\}}
 \big(
 \pseudoindex{-\nabla\psi\left(\mlc_{y_{i}}\mathbf{D}\mathbf{W}(t)^{\top}\mathbf{x}_{i}\right)}_{k} \nonumber 
 \\& \qquad \qquad \qquad  \qquad \qquad -
  t^{-1}\exp(-\tilde{\bfw}^\top \tilde{\bfx}_{i,k})  
 \mathbbm{1}_{\{i \in S_{k}\}} 
 \big)
 \tilde{\bfx}_{i,k}^\top \bfr(t) \nonumber 
 \\
 &\overset{(2)}{\le}   
 \sum^{N}_{i=1}
 \sum_{k \in [K] \setminus \{y_i\}}
 \big(
 \exp\left(
- \bfw(t)^\top \tilde{\bfx}_{i,k}
\right) \nonumber 
 \\& \qquad \qquad \qquad  \qquad \qquad -
  t^{-1}\exp(-\tilde{\bfw}^\top \tilde{\bfx}_{i,k})  
 \mathbbm{1}_{\{i \in S_{k}\}} 
 \big)
 \tilde{\bfx}_{i,k}^\top \bfr(t)
 \mathbbm{1}_{\{\tilde{\bfx}_{i,k}^\top \bfr(t) \ge 0\}} \nonumber 
 \\
 &\qquad + 
 \sum^{N}_{i=1}
 \sum_{k \in [K] \setminus \{y_i\}}
 \Big(
 \exp\left(
- \bfw(t)^\top \tilde{\bfx}_{i,k}
\right)
\big(1 - \sum_{k \in [K]} \exp(-\bfw(t)^\top \tilde{\bfx}_{i,k})\big) \nonumber  
 \\& \qquad \qquad \qquad \qquad  \qquad \qquad -
  t^{-1}\exp(-\tilde{\bfw}^\top \tilde{\bfx}_{i,k})  
 \mathbbm{1}_{\{i \in S_{k}\}} 
 \Big)
 \tilde{\bfx}_{i,k}^\top \bfr(t)  \mathbbm{1}_{\{\tilde{\bfx}_{i,k}^\top \bfr(t) < 0\}}. \label{eq: final_form_identical_soudry}
\end{align}

In (1) we used \Cref{lemma:pseudo-index-existence}, which implies that
\(
\tilde{\bfx}_{i,k}^\top \bfr(t)
=
\pseudoindex{\mlc_{y_{i}}\mathbf{D}\mathbf{R}(t)^{\top}\mathbf{x}_{i}}_{k}
\).
For (2), from the exponential tail upper/lower bound and \Cref{lemma:pseudo-label-exponential-tail-bounds}, we have that
\begin{align*}
\exp\left(
- \bfw(t)^\top \tilde{\bfx}_{i,k}
\right)
& \ge 
\pseudoindex{-\nabla\psi\left(\mlc_{y_{i}}\mathbf{D}\mathbf{W}(t)^{\top}\mathbf{x}_{i}\right)}_{k}
\\
& \ge
\exp\left(
- \bfw(t)^\top \tilde{\bfx}_{i,k}
\right)
\Big(1 - \sum_{k \in [K] \setminus \{y_{i}\}} \exp(-\bfw(t)^\top \tilde{\bfx}_{i,k})\Big).
\end{align*}

We note that Eqn. \eqref{eq: final_form_identical_soudry} above is identical to the right hand side of inequality (1) in \cite[Eqn.~(141)]{soudry2018implicit}.
Thus, the remainder of the analysis proceeds identically as in \citet[Lemma 20]{soudry2018implicit}.
\end{proof}

\section{A structural result on symmetric and convex functions}\label{section:convexity-structural-result}

\begin{proposition}\label{proposition:filling-the-gap}
  Let \(\psi : \bbR^{K-1}\to \bbR\)  be the template of a PERM loss that satisfies our Theorem 3.4.
Let \(\bfu^{t} \in \bbR^{K-1}\) be any sequence, where \(t=1,2,\dots\), such that
\[
\lim_{t\to \infty} \nabla \psi(\bfu^{t}) = \mathbf{0}
\]
is the zero vector.
Then
\(\lim_{t \to \infty}u^{t}_{j} = \infty\) for every \(j \in [K-1]\).
\end{proposition}
We prove \Cref{proposition:filling-the-gap} by first proving a structural result
(\Cref{theorem:boosting-vector-monotonicity-property})
 concerning symmetric and convex function \(f : \bbR^{n} \to \bbR\).
The proof of \Cref{proposition:filling-the-gap} will be presented in \Cref{section:application} as an application of the structural result, where we take \(f = \psi\), the template of a PERM loss, and \(n = K-1\), number of classes minus one.


Given a vector \(\bfx \in \bbR^{n}\) and a real number \(C \in \bbR\), define \(\bfx \vee C \in \bbR^{n}\) to be the vector such that
\[
  [\bfx \vee C]_{i} :=
  \max \{x_{i}, C \}, \quad \mbox{for all \(i \in [n]\)}.
\]
In other words, \(\bfx \vee C\) ``boosts'' entries of \(\bfx\) up to \(C\) if those entries are smaller than \(C\). Entries of \(\bfx\) larger than \(C\) are kept as-is.

Define \(\min(\bfx) = \min_{j\in[n]} x_{j}\) and  \(\mathrm{argmin}(\bfx) :=
\{ i \in [n] : x_{i} = \min(\bfx)\}
\).
We note the following easy-to-prove properties of the ``\(\vee\)'' operation:
\begin{enumerate}
  \item \(\min(\bfx \vee C) \ge C\) with equality if \(\min(\bfx) \le C\),

  \item \(\mathrm{argmin}(\bfx \vee C) \supseteq \mathrm{argmin(\bfx)}\).
\end{enumerate}

\begin{theorem}\label{theorem:boosting-vector-monotonicity-property}
  Suppose that \(f : \bbR^{n} \to \bbR\)
  is a symmetric, convex, and differentiable function.
  Then  for any real number \(C \in \bbR\) and any \(\bfx \in \bbR^{n}\),
  we have
  \[
    \tfrac{\partial f}{\partial x_{i}} ( \bfx )
    \le \tfrac{\partial f}{\partial x_{i}} (\bfx \vee C), \quad
    \mbox{
  for any
  \( i \in \mathrm{argmin}(\bfx)\)}.
  \]
\end{theorem}
Before proceeding with the proof (which is in \Cref{section:proof-of-the-structural-result}), we first introduce some necessary preliminary notations and facts.
Given a vector \(\bfx \in \bbR^{n}\), we define
\[
  \mathsf{val}(\bfx) := \{ x_{i} : i =1,\dots,n \}
\]
to be the \emph{set} of values consistings of the entries of \(\bfx\).
For example, if \(\bfx\) is the all-ones vector, then \(\mathsf{val}(\bfx) = \{1\}\).
Given \(v \in \mathsf{val}(\bfx)\), we let \(\mathsf{idx}(v, \bfx) = \{i \in \bfx : x_{i} = v\}\) be the set of indices that attains the value \(v\).

\textbf{Fact 1}: For a convex and differentiable function \(f : \bbR^{n} \to \bbR\), we have that
\begin{equation}
  \label{equation:monotonicity-property}
  \langle \nabla f(\bfx)  - \nabla f(\bfy) , \bfx - \bfy \rangle \ge 0,
  \quad \mbox{for all \(\bfx, \bfy \in \bbR^{n}\).}
\end{equation}
This is a simple and well-known consequence of convexity.
See \href{https://math.stackexchange.com/a/2820683}{this stackexchange answer} for a short proof.
When \(n=1\), Ineq.~\eqref{equation:monotonicity-property} is the fact that a convex differentiable function has nondecreasing derivative.
Ineq.~\eqref{equation:monotonicity-property} is also a consequence of \cite[Theorem 3.24]{phelps2009convex}.

\textbf{Fact 2}: For a symmetric  and differentiable function \(f : \bbR^{n} \to \bbR\), we have that
\begin{equation}
  \label{equation:equal-partial-derivatives}
    \tfrac{\partial f}{\partial x_{i}} ( \bfx )
  =
    \tfrac{\partial f}{\partial x_{j}} ( \bfx )
  ,\quad \mbox{whenever \(x_{i} = x_{j}\).}
\end{equation}
This fact follows from the chain rule and the definition of a symmetric function. To be precise, let \(\bfT : \mathbb{R}^n \to \mathbb{R}^n\) be the permutation matrix that switches the \(i\) and \(j\)-th coordinate. Then \(f (\bfx) = f(\bfT \bfx)\)
and moreover 
\(  \tfrac{\partial f}{\partial x_{i}} ( \bfx )
=
 \tfrac{\partial f}{\partial x_{i}} ( \bfT \bfx )
 =
  [\bfT \nabla f ( \bfT \bfx)]_i
  =
  [\nabla f ( \bfT \bfx)]_j
  =
  [\nabla f ( \bfx)]_j
  =
  \tfrac{\partial f}{\partial x_{j}} ( \bfx )
\).

\subsection{Proof of \Cref{theorem:boosting-vector-monotonicity-property}}\label{section:proof-of-the-structural-result}
Now, to prove the above theorem, we will use induction on ``\(m\)'' in the following lemma, which is simply a ``stratification'' of
\Cref{theorem:boosting-vector-monotonicity-property} into cases indexed by the ``parameter'' \(m\):
\begin{lemma}\label{lemma:boosting-vector-monotonicity-property-induction}
  Suppose that \(f : \bbR^{n} \to \bbR\)
  is a convex, symmetric, and differentiable function.
  Let \(m \in \{0,1,\dots, n\}\).
  Then  for any real number \(C \in \bbR\) and any \(\bfx \in \bbR^{n}\) with the property
  that  \( | \{ v \in \mathsf{val}(\bfx) : v < C\}| = m\),
  we have \[\tfrac{\partial f}{\partial x_{i}} ( \bfx ) \le \tfrac{\partial f}{\partial x_{i}} (\bfx \vee C), \quad
    \mbox{
  for any
  \( i \in \mathrm{argmin}(\bfx)\)}.
  \]
\end{lemma}
Note that if we have proved \Cref{lemma:boosting-vector-monotonicity-property-induction} for each \(m \in \{0,1,\dots, n\}\), then \Cref{theorem:boosting-vector-monotonicity-property} holds.

\textbf{The base step}:
we prove \Cref{lemma:boosting-vector-monotonicity-property-induction} when \(m = 0\) and \(m= 1\).
Strictly speaking, the proof-by-induction technique typically only involve \emph{only} the base case, which would be
the \(m=0\) case in this instance. But below, we will see that in the induction step, the \(m=1\) case is helpful.

Note that the \(m=0\) case holds vacuously, since \(\bfx \vee C = \bfx\). Below, we focus on the \(m=1\) case, where there exists a unique \(v \in \mathsf{val}(\bfx)\) such that \(v \le C\).
Let \(i \in \mathrm{argmin}(\bfx)\).
Note that we have \(\mathsf{idx}(v, \bfx) = \mathrm{argmin}(\bfx)\).
Using \Cref{equation:monotonicity-property} (Fact 1), we have that
\[
  \langle
  \nabla f (\bfx \vee C)
  -\nabla f(\bfx)
  ,
  (\bfx\vee C) - \bfx
  \rangle
  \ge 0.
\]
\begin{definition}
Given any set \(S \subseteq [n]\), we let \(\chi_{S} \in \bbR^{n}\) denote the \emph{characteristic vector on \(S\)}: \(\chi_{S}\) is the vector whose \(j\)th entry is \(=1\) if \(j \in S\) and \( = 0\) otherwise.

\end{definition}

By construction, we have
\[
  (\bfx\vee C) - \bfx
  =
  (C-v) \chi_{\mathsf{idx}(v, \bfx)}
  =
  (C-v) \chi_{\mathrm{argmin}(\bfx)}.
\]
The ``\(\tfrac{\partial f}{\partial x_{i}} (\cdot)\)'' notation for partial derivatives is a bit cumbersome. Instead, we will write ``\([\nabla f(\cdot)]_{i}\)'' from now on.
By \Cref{equation:equal-partial-derivatives} (Fact 2), we have
\[
[\nabla f(\bfx)]_{j}
=
[\nabla f(\bfx)]_{j'}, \quad \mbox{for all \(j,j' \in \mathsf{idx}(v, \bfx)\)}
\]
and likewise
\[
[\nabla f(\bfx \vee C)]_{j}
=
[\nabla f(\bfx \vee C)]_{j'}, \quad \mbox{for all \(j,j' \in \mathsf{idx}(v, \bfx)\)}.
\]

Thus, by \Cref{equation:equal-partial-derivatives}, we have
\[
  \langle
  \nabla f (\bfx \vee C)
  -\nabla f(\bfx)
  ,
  (\bfx\vee C) - \bfx
  \rangle
  =
  | \mathrm{argmin}(\bfx) |
  \cdot
  (C - v)
  ([\nabla f(\bfx \vee C)]_{i}
  -[\nabla f(\bfx )]_{i}).
\]
Now, since \(C > v\)
and \(
  | \mathrm{argmin}(\bfx) |
  > 0\), we must have that
\(
[\nabla f(\bfx \vee C)]_{i}
  -[\nabla f(\bfx )]_{i}
  \ge 0
\), as desired. This proves the base step.

\textbf{Induction step}: Suppose \Cref{lemma:boosting-vector-monotonicity-property-induction} holds for every integer \(m\) where \(0 \le m < n\), we must show that \Cref{lemma:boosting-vector-monotonicity-property-induction} also holds for \(m + 1\).
To this end, let \(\bfx \in \bbR^{n}\) and \(C \in \bbR\)  be such that
\( | \{ v \in \mathsf{val}(\bfx) : v < C\}| = m+1\).
Let \(v_{1},\dots, v_{m+1} \in \bbR\) be all the elements of \(\{v \in \mathsf{val}(\bfx) \mid v < C\}\) enumerated in increasing order,
i.e.,
\(v_{1} < \cdots < v_{m+1}\).

Note by construction, we have that
\(\{v \in \mathsf{val}(\bfx) \mid v < v_{m+1}\} = \{v_{1},\dots, v_{m}\}\) and so we immediately get that
\(| \{v \in \mathsf{val}(\bfx) \mid v < v_{m+1}\}| = m\).
By the \(m\)-th case of \Cref{lemma:boosting-vector-monotonicity-property-induction}  (i.e., the induction hypothesis)
using \( v_{m+1}\) as \(C\),
we get
\begin{equation}
  \label{equation:induction-step-useful-intermediary-step}
  [
 \nabla f( \bfx \vee v_{m+1} )
  ]_{i}
 \ge
  [
 \nabla f( \bfx  )
 ]_{i}
 \quad \mbox{for any \(i \in \mathrm{argmin}(\bfx)\).}
\end{equation}
\textcolor{black}{Below fix some \(i \in \mathrm{argmin(\bfx)}\)} arbitrarily.
Let \(\bfx' := \bfx \vee v_{m+1}\). \textcolor{black}{We note that by construction, all the entries of \(\bfx'\) that are less than \(C\) are set to equal to \(v_{m+1}\).} In other words,  \[ \{ v \in \mathsf{val}(\bfx') : v < C\} = \{v_{m+1}\}\] is a singleton set.
Thus, by the \(m=1\) case of \Cref{lemma:boosting-vector-monotonicity-property-induction} \textcolor{black}{applied to \(\bfx'\)}, we get that
\[
  [\nabla f(\bfx' \vee C)]_{i'}
  \ge
  [\nabla f(\bfx' )]_{i'},
  \quad \mbox{for any \(i' \in \mathrm{argmin}(\bfx')\).}
\]
Since
\(
\mathrm{argmin}(\bfx') \supseteq
\mathrm{argmin}(\bfx)
\), \textcolor{black}{we have that \(i \in \mathrm{argmin}(\bfx')\) as well (recall that \(i\) was chosen earlier from \(\mathrm{argmin}(\bfx)\) arbitrarily). Thus the above inequality implies in particular that}
\[
  [\nabla f(\bfx' \vee C)]_{i}
  \ge
  [\nabla f(\bfx' )]_{i}
  =
  [\nabla f(\bfx\vee v_{m+1} )]_{i}.
\]
Combined with \Cref{equation:induction-step-useful-intermediary-step}, we get
\[
  [\nabla f(\bfx' \vee C)]_{i}
  \ge
  [\nabla f(\bfx )]_{i}.
\]
\textcolor{black}{Finally, we note that \(\bfx' \vee C = (\bfx \vee v_{m+1}) \vee C = \bfx \vee C \).} Thus, the above implies
\[
  [
 \nabla f( \bfx \vee C )
  ]_{i}
 \ge
  [
 \nabla f( \bfx  )
 ]_{i}
 \quad \mbox{for any \(i \in \mathrm{argmin}(\bfx)\)}
\]
since the choice of \(i \in \mathrm{argmin}(\bfx)\) was arbitrary.

\subsection{Application to our setting}\label{section:application}
The condition
\(\lim_{t \to \infty}u^{t}_{j} = \infty\) by definition
means that for every real number \(M \in \bbR\), there exists \(T\) such that for all \(t \ge T\) we have
\(u^{t}_{j} > M\).
Thus, suppose that there exists \(j \in [K-1]\) such that
\(\lim_{t \to \infty}u^{t}_{j} \ne \infty\), then there exists a real number \(M \in \bbR\) such that for all \(T=1,2,\dots\) there exists some \(t \ge T\) such that \(u^{t}_{j} \le M\).
Passing to a subsequence, we assume that \(u^{t}_{j} \le M\) (and so \(\min(\bfu^{t}) \le M\)) for all \(t=1,2,\dots\).
Note that \(\lim_{t \to \infty} \nabla \psi(\bfu^{t}) = \mathbf{0}\) continues to hold.

Below, whenever we say ``for all/every \(t\)'', we mean ``for all/every \(t=1,2,\dots\)''.

Onto the proof. First recall the lower bound portion of the exponential tail property:
\begin{align}
  &\mbox{For all \(\bfu \in \bbR^{K-1}\) such that } \min(\bfu) > u_-,
  \,\, \mbox{we have}
  \nonumber
  \\
  &
  -[
  \nabla \psi(\bfu)
  ]_i
  \ge
  c \Big(1 - \sum_{j=1}^{K-1} \exp(-u_j)\Big) \exp(-u_i),\quad \mbox{for all \(i \in [K-1]\)}. \label{equation:exponential-tail-lower-bound}
\end{align}
Let
\(  C :=
  \max
  \{\,\,
  2|u_{-}|, \,\, M,\,\, -\log (\tfrac{1}{2(K-1)}) \,\,
  \}
\)
and
\(\bfv^{t} := \bfu^{t} \vee C\) for all \(t\).
This choice of \(C\) (and \(\bfv^{t}\)) has the following consequences:
\begin{enumerate}

  \item The fact that
        \(C \ge -\log (\tfrac{1}{2(K-1)})\)
        implies
       \(\Big(1 - \sum_{j=1}^{K-1} \exp(-v_j)\Big) \ge \frac{1}{2}\).
  \item
If \(\bfv \in \bbR^{K-1}\) is such that \(\min(\bfv) \ge C\), then we have by \Cref{equation:exponential-tail-lower-bound} that
\begin{equation}
  -[
  \nabla \psi(\bfv)
  ]_i
  \ge
  \frac{1}{2}
  c  \exp(-v_i).
  \label{equation:consequence-of-ET}
\end{equation}

  \item \(\min(\bfu^{t}) \le C\). This is true since \(\min(\bfu^{t}) \le M\).

  \item
Choose \(i_{t} \in \mathrm{argmin}(\bfu^{t}) \) for every \(t\). Then \(v^{t}_{i_{t}} = \min(\bfv^{t})= C\) for every \(t\). This is simply a consequence of the fact that
        \(
\mathrm{argmin}(\bfu^{t})
        \subseteq
\mathrm{argmin}(\bfv^{t})
        \).

  \item
 We have
\(  \lim_{t \to \infty} [\nabla \psi(\bfu^{t})]_{i_{t}} = 0
\).
        This follows from the assumption that
\(  \lim_{t \to \infty} \nabla \psi(\bfu^{t})= \mathbf{0}
\).
\end{enumerate}

By \Cref{theorem:boosting-vector-monotonicity-property}, we have for every \(t\) that
  \[
    [\nabla \psi(\bfu^{t})]_{i_{t}}
    \le
    [\nabla \psi(\bfv^{t})]_{i_{t}}.
  \]
  By plugging in \(\bfv^{t}\) for \(\bfu\) in \Cref{equation:consequence-of-ET} above and the fact that \(v^{t}_{i_{t}} = C\), we have
  \[
    [\nabla \psi(\bfv^{t})]_{i_{t}}
    \le
    -
  \tfrac{1}{2}
  c  \exp(-v_{i_{t}})
  =
    -
  \tfrac{1}{2}
  c  \exp(-C)
  < 0.
  \]
  Since \(
    -
  \tfrac{1}{2}
  c  \exp(-C)
  \)
  is a constant that doesn't depend on \(t\), it is impossible for
  \(    \lim_{t\to\infty}
    [\nabla \psi(\bfv^{t})]_{i_{t}}
    =0
\)
  to hold. This proves \Cref{proposition:filling-the-gap}.

\section{On the existence of \(\tilde{\mathbf{w}}\)}\label{section:wtildeconjecture}


The goal of this section is to explain the challenge and the current gap in the proof of  the existence of \(\tilde{\bfw}\) that satisfies the condition in \Cref{equation:w-tilde-defining-condition} for almost all linearly separable datasets.
To this end,
recall \Cref{eq: multiclass_SVM}, the hard-margin SVM formulated as a constrained optimization:
\begin{equation}
\label{equation:hard-margin-SVM-primal-problem-appendix}
\hat{\mathbf{w}} = \argmin_{\mathbf{w}} \frac12 \Vert \mathbf{w}\Vert^2 \,\text{ s.t.}\,\forall n,\forall k \ne y_n : \mathbf{w}_{y_n}^\top\mathbf{x}_n \ge \mathbf{w}_k^\top\mathbf{x}_n + 1.
\end{equation}
Moreover, recall that \(\mathcal{S}_k\), the set of support vectors for each \(k \in [K]\), is defined by 
\[\mathcal{S}_k :=
\{n :(\hat{\mathbf{w}}_{y_n}-\hat{\mathbf{w}}_k)^\top\mathbf{x}_n  =1\}.
\]
The Lagrangian of the objective in
\Cref{equation:hard-margin-SVM-primal-problem-appendix}
is
\begin{equation}
    \label{equation:hard-margin-SVM-lagrangian}
L(\bfw, \boldsymbol{\alpha}) =\frac{1}{2}\sum_{r=1}^{K}\left\Vert \mathbf{w}_{r}\right\Vert ^{2}+\sum_{n=1}^{N}\sum_{r\neq y_{n}}\alpha_{n,r}\left(\mathbf{w}_{y_{n}}-\mathbf{w}_{r}\right)^{\top}\mathbf{x}_{n}
\end{equation}
where $\alpha_{n,r}$ are the dual variables.
Let \(\delta_{i,j}\) denote the Kronecker delta, i.e., \(\delta_{i,j} = 1\) if \(i=j\) and \(\delta_{i,j} = 0\) otherwise.
Taking the gradient of \(L(\bfw, \boldsymbol{\alpha})\) with respect to $\mathbf{w}_{k}$, we get
\[
\mathbf{w}_{k}+\sum_{n=1}^{N}\sum_{r\neq y_{n}}\alpha_{n,k}\left(\delta_{r,y_{n}}-\delta_{r,k}\right)\mathbf{x}_{n}=\mathbf{w}_{k}+\sum_{n=1}^{N}\left(\delta_{k,y_{n}}\sum_{r\neq y_{n}}\alpha_{n,r}-\alpha_{n,k}\right)\mathbf{x}_{n}.
\]
So the KKT conditions satisfied by a stationary point \(\hat{\bfw}\) (hence globally optimal for \Cref{equation:hard-margin-SVM-primal-problem-appendix}) are
\begin{align}
& \forall k\in\left[K\right]: \hat{\mathbf{w}}_{k}=\sum_{n=1}^{N}\left(\alpha_{n,k}-\delta_{k,y_{n}}\sum_{r\neq k}\alpha_{n,r}\right)\mathbf{x}_{n}\label{eq:proof-of-w-existence-KKT-condition}\\
& \forall k\in\left[K\right]:\forall n: 
\mbox{one of the following holds}
\begin{cases}
\alpha_{n,k}\geq0\,\,\,\mathrm{and}\,\,\,\left(\hat{\mathbf{w}}_{y_{n}}-\hat{\mathbf{w}}_{k}\right)^{\top}\mathbf{x}_{n}=1\,\, \\\alpha_{n,k}=0\,\,\,\mathrm{and}\,\,\,\left(\hat{\mathbf{w}}_{y_{n}}-\hat{\mathbf{w}}_{k}\right)^{\top}\mathbf{x}_{n}>1\,\label{eq:proof-of-w-existence-KKT-condition-complementary-slackness}
\end{cases}
\end{align}
where  Eqn.~\eqref{eq:proof-of-w-existence-KKT-condition-complementary-slackness}  (the second line) above is the complementary slackness condition.

The goal of this section is to prove the following result regarding the existence of \(\tilde{\bfw}\) that satisfies the condition in \Cref{equation:w-tilde-defining-condition}, which we restate below:
\begin{equation}
\label{equation:w-tilde-defining-condition-appendix}
\forall k \in [K], \forall n \in \mathcal{S}_k:\,\eta\exp\left(-\mathbf{x}_{n}^{\top}\left(\tilde{\mathbf{w}}_{y_n}-\tilde{\mathbf{w}}_{k}\right)\right)=\alpha_{n, k}.
\end{equation}

\begin{conj}\label{proposition:w-tilde-existence}
    For almost all  linearly separable multiclass datasets, Assumption~\ref{assumption:w_tilde_existence} holds, i.e., 
    Eqn.~\eqref{equation:w-tilde-defining-condition-appendix}
has a solution \(\tilde{\mathbf{w}}\).
\end{conj}
Below, we use the word ``generically'' to mean ``for linearly separable datasets outside of a set of Lebesgue measure zero''.
In order for \eqref{equation:w-tilde-defining-condition-appendix} to have a solution in \(\tilde{\mathbf{w}}\) generically, two conditions need to hold (generically).

\noindent \textbf{Condition 1.}
 \(\alpha_{n,k} >0\) for all
\(k\) and \(n\) such that
\(
n \in \mathcal{S}_k :=
\{n :(\hat{\mathbf{w}}_{y_n}-\hat{\mathbf{w}}_k)^\top\mathbf{x}_n  =1\}
\). 

Condition 1 is already nontrivial and a gap in proving the Conjecture, as we will see below. For the sake of explaining Condition 2 below, let us assume  Condition 1 holds.  Then we can rewrite
\eqref{equation:w-tilde-defining-condition-appendix} as

\begin{equation}
\label{equation:w-tilde-defining-condition-appendix-2}
\forall k \in [K], \forall n \in \mathcal{S}_k:\, \mathbf{x}_{n}^{\top}\left(\tilde{\mathbf{w}}_{y_n}-\tilde{\mathbf{w}}_{k}\right)= \log\left(\frac{\eta}{\alpha_{n, k}}\right).
\end{equation}

Define the vector \(\mathbf{m}_{n,k}\) obtained by taking the difference between the \(k\)-th and \(y_n\)-th elementary basis vector in \(\bbR^{K}\), i.e.,
\begin{align*}
\mathbf{m}_{n,k} & :=\mathbf{e}_{k}-\mathbf{e}_{y_{n}} \in \bbR^{K}.
\end{align*}
Then we can further rewrite
\eqref{equation:w-tilde-defining-condition-appendix-2}
as
\begin{equation}
\label{equation:w-tilde-defining-condition-appendix-3}
\forall k \in [K], \forall n \in \mathcal{S}_k:\, (\mathbf{m}_{n,k} \otimes \mathbf{x}_{n})^{\top}
\tilde{\mathbf{w}}
= \log\left(\frac{\eta}{\alpha_{n, k}}\right).
\end{equation}
It is more convenient to pool all the class-specific support vectors \(\mathcal{S}_k\) into a single set: $\mathcal{S}\triangleq\left\{ \left(n,k\right):\left(\hat{\mathbf{w}}_{y_{n}}-\hat{\mathbf{w}}_{k}\right)^{\top}\mathbf{x}_{n}=1\right\} $. For readability, we linearly order the tuples in \(\mathcal{S}\), i.e., we assign to each $\left(n,k\right)\in\mathcal{S}$  a unique index $i\in\left\{ 1,...,\left|\mathcal{S}\right|\right\} $. 
In other words, we define \(n(1),\dots, n(|\mathcal{S}|)\) 
and \(k(1),\dots, k(|\mathcal{S}|)\)  such that
\[
\mathcal{S} = \{
(n(1), k(1)) , \, (n(2), k(2)), \, \dots, \, (n(|\mathcal{S}|),
k(|\mathcal{S}|))
\}.
\]
To reduce notational clutter in the subscript, define $\olbfx_{i}\triangleq\mathbf{x}_{n(i)}$
and
$\olbfm_{i}\triangleq\mathbf{m}_{n(i),k(i)}$.
Finally, define
\[
\olbfM \triangleq\left[\olbfm_{1},\dots,\olbfm_{\left|\mathcal{S}\right|}\right]
\mathbb{R}^{K\times\left|\mathcal{S}\right|}
,\,\, \olbfX\triangleq\left[\olbfx_{1},\dots,\olbfx_{\left|\mathcal{S}\right|}\right]
\in\mathbb{R}^{d\times\left|\mathcal{S}\right|},\,\, \mbox{and}
\,\,
\mathbf{G}\triangleq\left(\mathbf{M}\circ\mathbf{X}\right)\in\mathbb{R}^{dK\times\left|\mathcal{S}\right|}.
\]
with $\circ$ denoting the Khatri-Rao product, which is, by definition, the matrix obtained by taking the Kronecker product of corresponding columns \citep{khatri1968solutions}. Note that the Khatri-Rao product is only defined for two matrices that have the same number of columns. See \cite{liu1999matrix} for a  reference. We now state

\noindent \textbf{Condition 2.}
\(\rank(\mathbf{G}) = | \mathcal{S}|\) generically. 

Note that given Condition 2, 
Eqn.~\eqref{equation:w-tilde-defining-condition-appendix-3}
has a solution in \(\tilde{\mathbf{w}}\), while Condition 1 is necessary for the logarithm in \eqref{equation:w-tilde-defining-condition-appendix-3}
to be valid in the first place.

The challenge in proving Condition 2 in the multiclass case is that the column vectors of \(\olbfX\) may have repeats, i.e., it is possible for \(n(i) = n(i')\) when \(i \ne i'\). 
It is easy to generate synthetic linearly separable multiclass datasets satisfying this condition.
Nonetheless, we observe that even in such a case, the matrix \(\mathbf{G}\) has rank \(|\mathcal{S}|\), i.e., Condition 2 holds. We verify this experimentally in the Python notebook \texttt{checking\_conjecture\_in\_Appendix\_H.ipynb} available at

\url{https://github.com/YutongWangML/neurips2024-multiclass-IR-figures}

In the binary case, linear classifiers are parametrized simply as a single vector, rather than the more cumbersome one-vector-per-class parametrization.
Under the one-vector parametrization, the \(\olbfM\) matrix becomes a \(1\)-by-\(|\mathcal{S}|\) matrix consistings of only \(\pm 1\)'s, and \(\mathbf{G} \)  reduces to \(\olbfX\). Moreover \(\olbfX\) has no repeats. 
Thus, Condition 2 holds trivially.
In both the multiclass and binary settings, given Condition 2, the proof for Condition 1 can proceed exactly as in Lemma 12 from \cite{soudry2018implicit} where their \(\mathbf{X}_{\mathcal{S}}\) is replaced by our \(\mathbf{G}\).

\section{Additional Experiments}
\label{section:experiments}
We provide additional experimental support for our main theoretical result for the PairLogLoss \citep{wang2022rank4class}.
Code for recreating the figures can be found at 

\url{https://github.com/YutongWangML/neurips2024-multiclass-IR-figures}

The code can be ran on Google Colab with a CPU runtime in under one hour.

\begin{figure}
    \centering
        \includegraphics[width=0.32\linewidth]{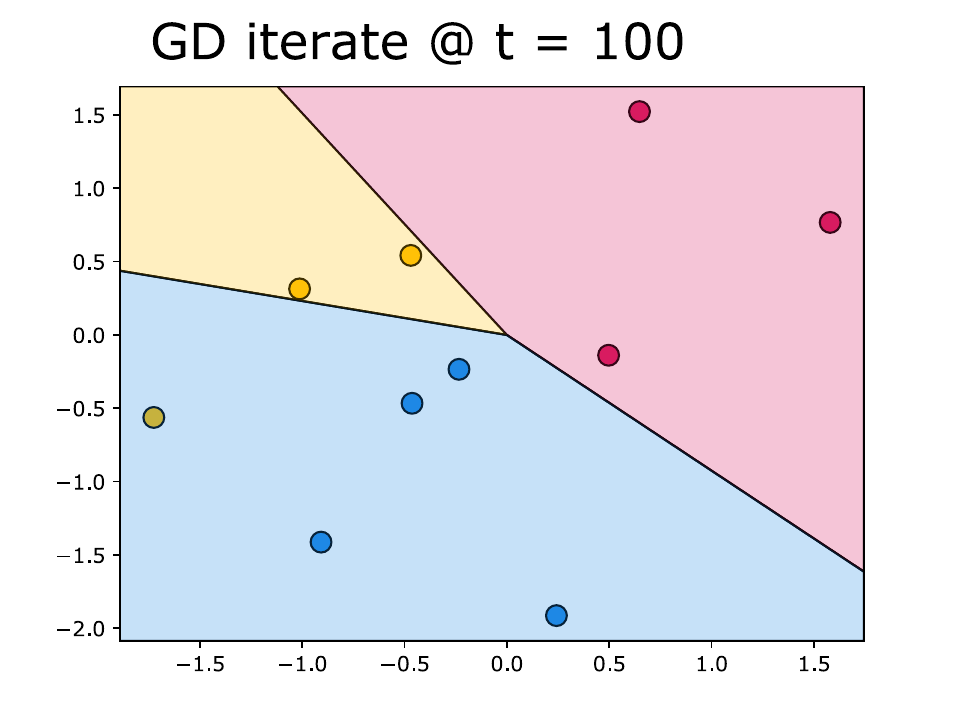}
\includegraphics[width=0.32\linewidth]{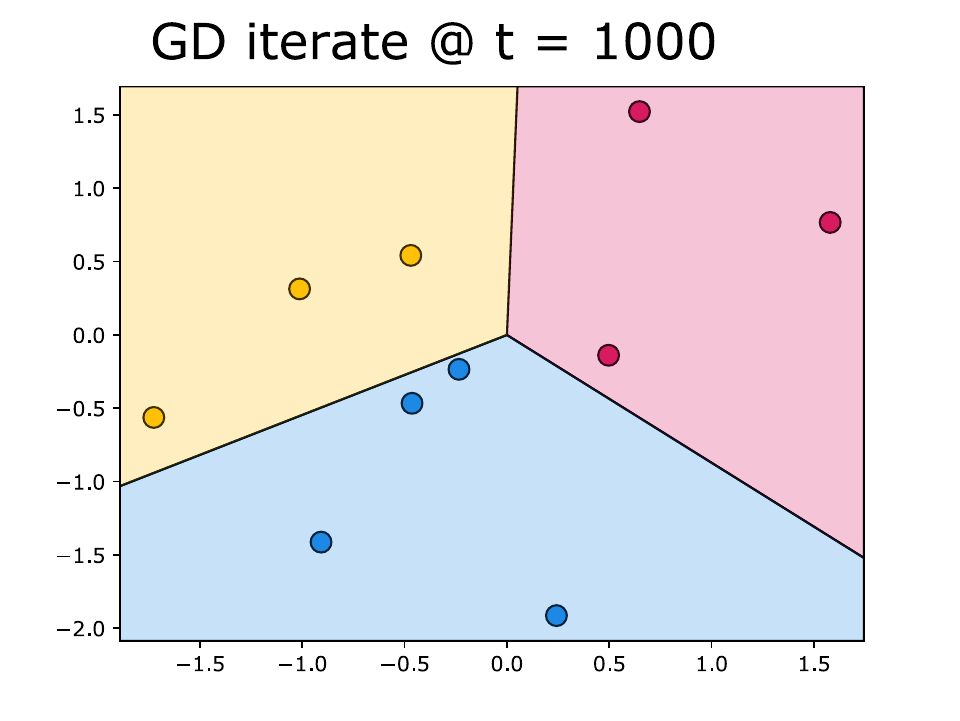}
\includegraphics[width=0.32\linewidth]{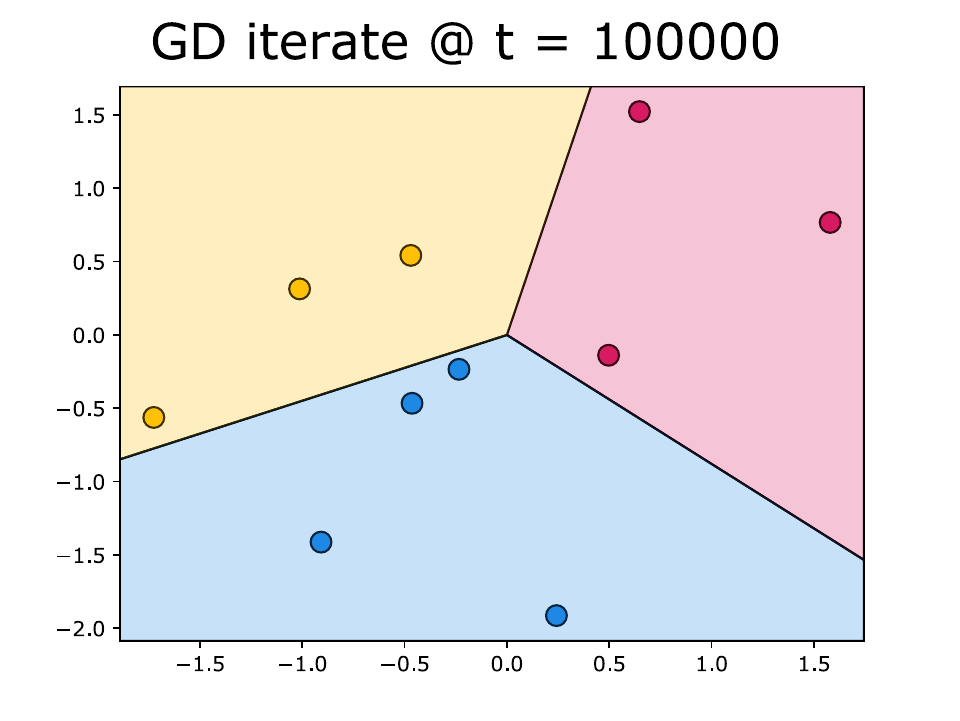}
    ~
\includegraphics[width=0.32\linewidth]{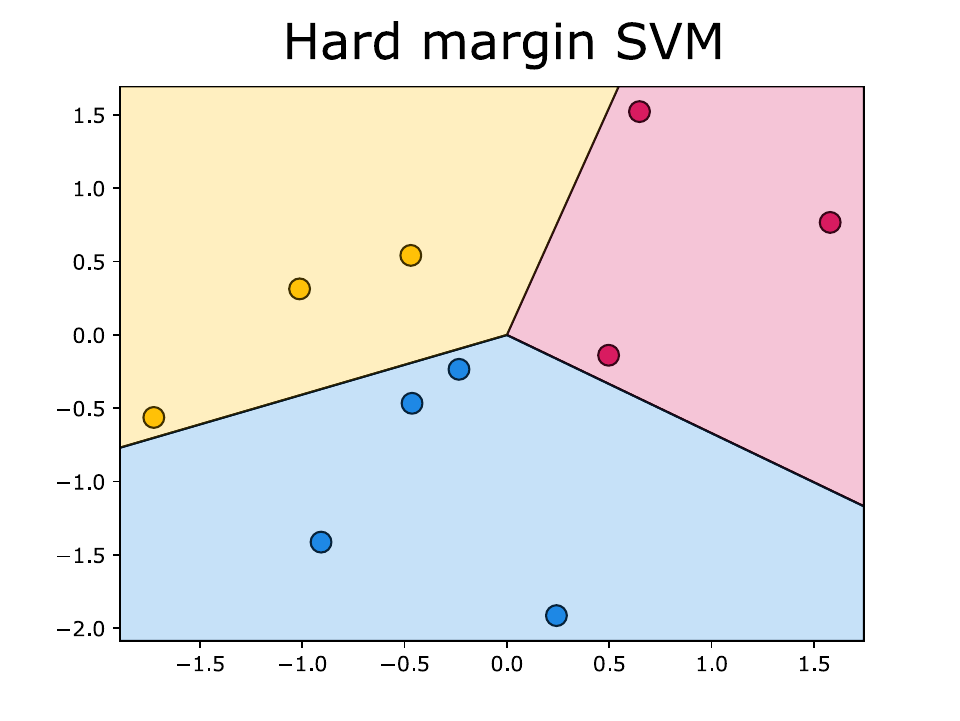}
    \caption{
    Small simulation with \(N=10\), \(d = 2\) and \(K=3\).
    The loss used is the ``PairLogLoss''.
    \emph{Top row.} Decision regions of classifiers along the gradient path \(\bfw(t)\) at \(t = 100,\,1000\), and \(100000\), respectively from left to right.
    \emph{Bottom row.} Decision regions of the hard-margin multiclass SVM.
    Note that most of the progress is made between iterations 100 and 1000.
    }
    \label{fig:small-scale-experiment}
\end{figure}

\begin{figure}
    \centering
    \includegraphics[width=1.0\linewidth]{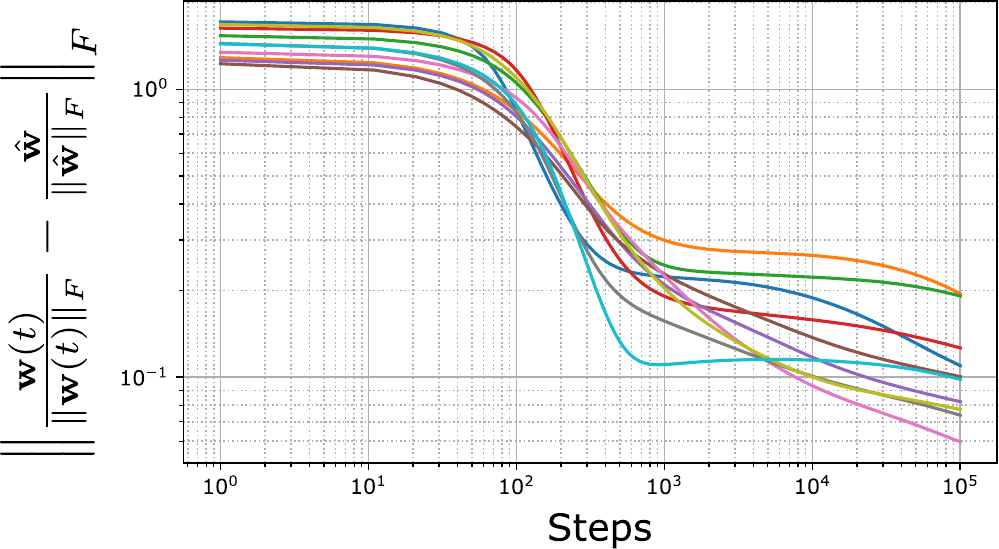}
    \caption{
        Large simulations with \(N=100\), \(d = 10\) and \(K=3\). The loss used is the ``PairLogLoss''. The curves are 10 independent runs with randomly sampled data and random initialization for gradient descent over \(100000\) iterations.
Note that the convergence in direction of the gradient descent iterates to the hard-margin SVM slows down in log-log space.
    }
    \label{fig:large-scale-experiment}
\end{figure}

\end{document}